\documentclass[preprint, 5p]{elsarticle}

\usepackage{hyperref, amsmath, amssymb, amsfonts, graphicx, subcaption, xcolor, hyperref, wrapfig, tabu, booktabs, siunitx, caption, float}

\journal{Applied Intelligence}

\newcommand{\R}{\mathbb{R}}
\newcommand{\E}{\mathrm{E}}

\renewcommand{\Pr}{p}

\newcommand{\vis}{\textrm{vis}}
\newcommand{\aud}{\textrm{aud}}
\newcommand{\mov}{\textrm{mov}}

\graphicspath{{images}}

\begin{document}

\begin{frontmatter}

\title{Variational Meta Reinforcement Learning for Social Robotics}

\author{Anand Ballou\corref{cor1}}
\cortext[cor1]{Corresponding author}
\ead{anand.ballou@inria.fr}

\author{Xavier Alameda-Pineda}
\ead{xavier.alameda-pineda@inria.fr}

\author{Chris Reinke}
\ead{chris.reinke@inria.fr}

\address{Inria, Univ. Grenoble Alpes,  CNRS, Grenoble INP, LJK}
\address{655, Avenue de l'Europe, 38334, Montbonnot, France}




\begin{abstract}
    With the increasing presence of robots in our everyday environments, improving their social skills is of utmost importance.
    Nonetheless, social robotics still faces many challenges.
    One bottleneck is that robotic behaviors often need to be adapted, as social norms depend strongly on the environment.
    For example, a robot should navigate more carefully around patients in a hospital than around workers in an office.
    In this work, we investigate meta-reinforcement learning (meta-RL) as a potential solution.
    Here, robot behaviors are learned via reinforcement learning, where a reward function needs to be chosen so that the robot learns an appropriate behavior for a given environment.
    We propose to use a variational meta-RL procedure that quickly adapts the robots' behavior to new reward functions.
    As a result, in a new environment different reward functions can be quickly evaluated and an appropriate one selected.
    The procedure learns a vectorized representation for reward functions and a meta-policy that can be conditioned to such a representation.
    Given observations from a new reward function, the procedure identifies its representation and conditions the meta-policy to it.
    While investigating the procedure's capabilities, we realized that it suffers from posterior collapse where only a subset of the dimensions in the representation encode useful information, resulting in reduced performance.
    Our second contribution, a radial basis function (RBF) layer, partially mitigates this negative effect.
    The RBF layer lifts the representation to a higher dimensional space, which is more easily exploitable for the meta-policy. 
    We demonstrate the interest of the RBF layer and the usage of meta-RL for social robotics in four robotic simulation tasks.
\end{abstract}


\end{frontmatter}

\section{Introduction}

\begin{figure}
\centering
\includegraphics[width=\columnwidth]{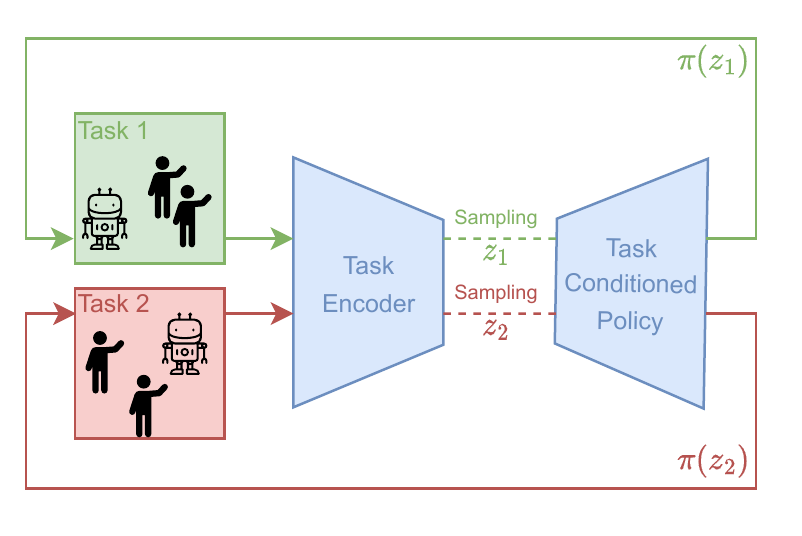}
\caption{ \textbf{Variational Meta-RL for social robotics.}
 Meta-RL enables robots to quickly adapt to task changes using only a few observations of the target task, making it an efficient approach for training social behaviors in robots. 
 We use a variational encoder network to convert task observations into a task representation in the form of a vector. 
 The meta-policy, which is the robot's behavior, is then conditioned on this task representation. 
 The robot can quickly adapt to a new task by collecting a small number of observations and computing the corresponding task representation. 
 This representation is then used to condition the meta-policy for the new task.}
\label{fig:teaser}
\end{figure}

Since the emergence of social robotics~\cite{fong2003survey}, substantial efforts have been made to enable the deployment of interactive robots in different environments, such as industrial~\cite{liu2022asymmetric}, healthcare~\cite{davison2020working}, or education~\cite{kubota2020jessie} facilities.
However, this deployment is very challenging, since socially appropriate behaviors are strongly context-dependent, making it necessary to adapt them to each target environment.
For example, a social robot might be used to navigate a care center for the elderly or an office.
In the care center, the robot should avoid getting too close to people to ensure that they do not feel unsafe.
Conversely, in an office environment this restriction might be less important and the robot may move closer to people to reach its target position faster.
Being able to adapt a robot's behavior quickly to the specific needs of a new environment is essential to the practical application of social robotics. 

Given the complex dynamics of social environments, the approach of manually designing and adapting behaviors for each environment, for example by finite state machines~\cite{colledanchise2018behavior}, is often not feasible. 
As an alternative, reinforcement learning (RL) \cite{sutton2018reinforcement} can be used for training social behaviors.
However, classical RL introduces its own set of problems when applied to social robotics~\cite{akalin2021reinforcement}.
The most prominent problem is that RL requires large amounts of training data, meaning hours of observations that have to be collected while the robot is learning behaviors for a new environment.

We propose the use of meta-RL~\cite{beck2023survey} to overcome this issue (Fig.~\ref{fig:teaser}). 
Meta-RL aims to adapt the learning process to a certain problem domain so that it can efficiently solve a new target task in this domain.
Of particular interest to us are methods that are able to adapt using only a few observations of the target task.
In our case, tasks represent the different environments of a social robot.  
Each task is defined by a reward function that defines which states a robot should try to reach or avoid.
It is often a weighted sum over several reward components: $R(s) = \sum_i w_i r_i(s)$.
For example, for social navigation \cite{chen2017decentralized,zhou2022robot}, the reward function might include a positive component for reaching the robot's target position and a negative component for getting too close to people.
Depending on the environment, such as in our office vs.\ care center example, we choose a different reward function for each environment.
In a care center, the reward function might have a stronger negative component to prevent the robot from getting too close to patients.
Conversely, in an office environment, the positive component for reaching the goal might be stronger so that the robot reaches its destination faster.
Meta-RL aims to adapt quickly to such task changes.

Even though meta-RL should efficiently adapt to new tasks (i.e.\ reward functions), to our knowledge its use in social robotics has scarcely been explored. 
To date, the only study on the topic uses on-policy meta-RL, requiring lots of interactions with the environment before reaching a satisfactory performance~\cite{lie2020fficient}. 
To cope with this issue, we propose to explore the use of off-policy meta-RL methods for social robotics tasks.

More precisely, we will focus on meta-RL methods that learn task-conditioned policies, since they are able to adapt using only a few observations (or environment steps). 
These methods learn a meta-policy, i.e., a behavior that is conditioned to a task representation in the form of a vector $z \in \mathbb{R}^d$.
An encoder network that takes input observations of a task is used to compute $z$.
The encoder network and the meta-policy are learned concurrently in an end-to-end way on a set of random source tasks.
After this meta-training, given a new target task, a small number of observations are collected to compute its task representation and to condition the meta-policy to it.

\begin{figure}[t]
\centering
\begin{minipage}[b]{.27\columnwidth}
\includegraphics[width=\textwidth]{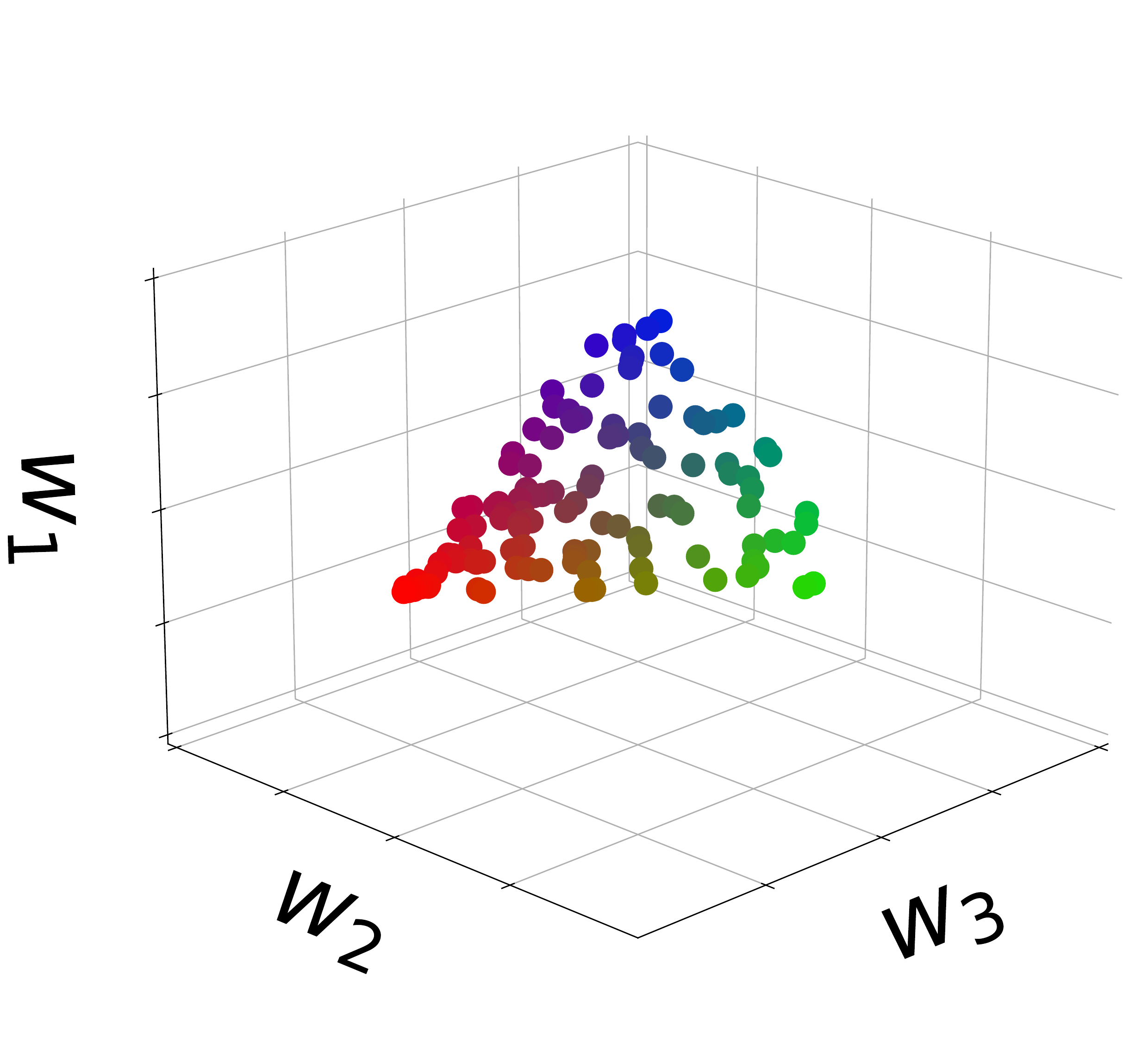}
\includegraphics[width=\textwidth]{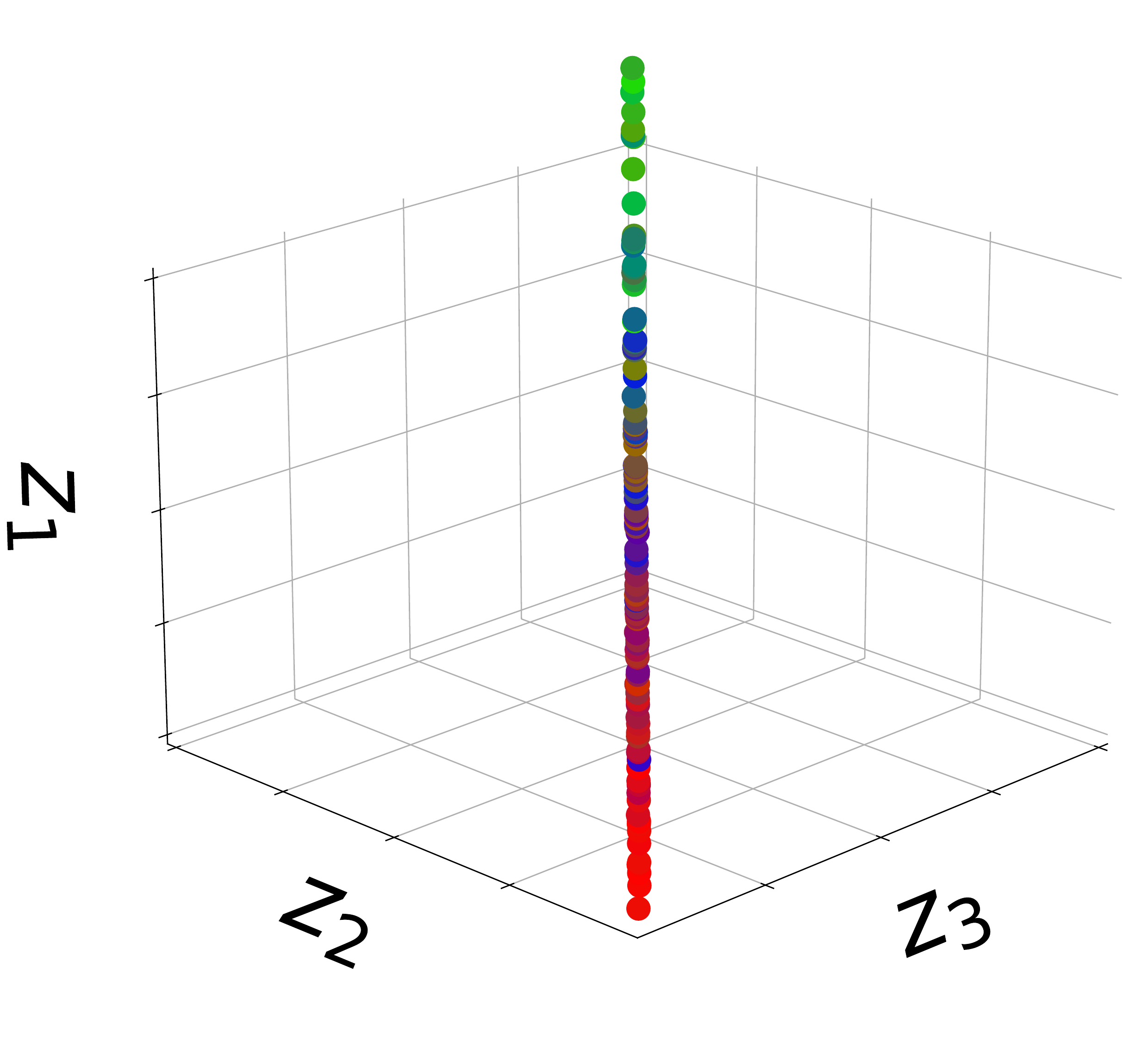}
\end{minipage}
\includegraphics[width=.68\columnwidth]{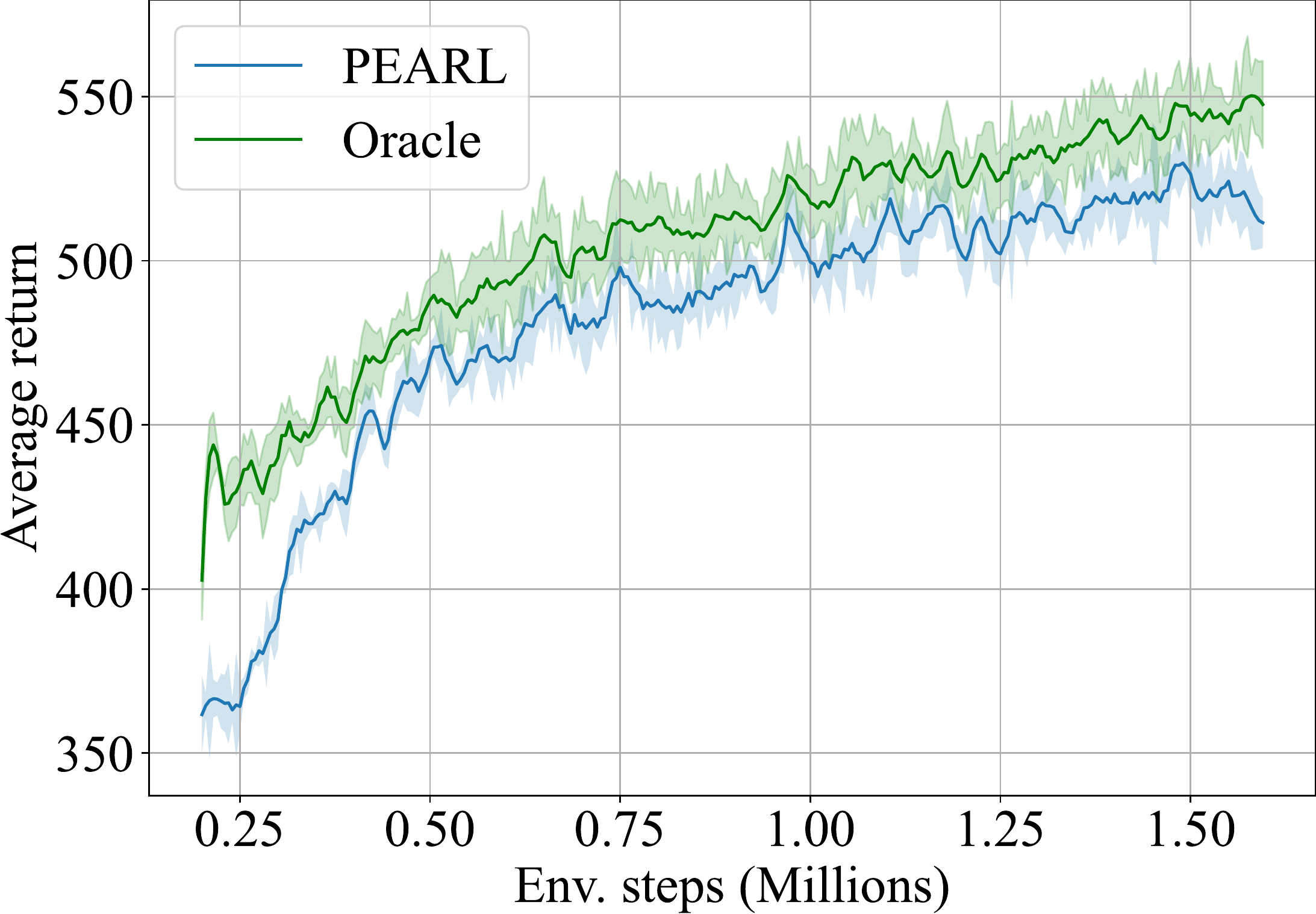}
\caption{
\textbf{Posterior collapse in PEARL.}
(left-top) The ground truth task representation, where each task corresponds to a colored dot. 
The three primary colors correspond to the strength of each of the three reward component weights.
(left-bottom) PEARL's learned task representation uses only one dimension ($z_1$) and exhibits posterior collapse in the two other dimensions ($z_2$ and $z_3$).  
(right) As a result, PEARL's learning performance compared to a task-conditioned policy using the ground truth (Oracle) is reduced.
The proposed RBF-PEARL aims to close this gap by mitigating the negative impact of posterior collapse during training.
}
\label{fig:posterior_collapse}
\end{figure}

A promising research direction for task-conditioned policies is variational architectures~\cite{kingma2013auto}, such as PEARL~\cite{rakelly2019efficient}.
Instead of learning a deterministic representation of tasks, PEARL learns to represent them via a distribution, exploiting the flexibility of probabilistic models and the representation power of deep neural networks~\cite{liu2021ngdnet,girin2021dynamical}. 
We investigated the usability of such variational meta-RL frameworks for social robotic tasks and realized that the learned encoder suffers from posterior collapse resulting in reduced learning performance (Fig.~\ref{fig:posterior_collapse}).
We propose to use a radial basis function (RBF) layer \cite{broomhead1988radial} to transform the task representation before giving it to the downstream task-conditioned policy, thus constructing an embedding that is more suited to representing tasks. 
RBF networks are universal function approximators~\cite{park1991universal} whose parameters can be learned and exhibit interesting results in classification tasks~\cite{vidnerova2018deep, abpeykar2019ensemble} as well as in value-learning for continuous action DRL~\cite{asadi2020deep}. 

In summary, our contribution is twofold:
\begin{enumerate}
    \item We successfully demonstrate the use of off-policy meta-RL on three robotics tasks and four different settings by quickly adapting to various reward functions.
    \item We improve the existing PEARL algorithm by introducing an RBF layer that transforms the task representation, enabling better training of the task-dependent behavior. 
\end{enumerate}

In this paper, we first discuss other work related to our topic, then introduce our methodology, and finally present the experimental evaluation.

\section{Related work}

\subsection{Adaptive Social Robotics and Reinforcement Learning}

Social robotics aims to design robots that share the same space as humans and interact with them in a natural and interpersonal manner.
This includes tasks such as human-aware navigation~\cite{moller2021survey} or controlling a robot's gaze to perceive humans optimally \cite{lathuiliere2019neural}.
A general overview of social robotics, including methods and their potential application domains, can be found in the following surveys: \cite{breazeal2016social, sheridan2020review, henschel2021makes}.
An important factor for the successful deployment of social robots is their adaptability to different users and social environments.
The following surveys provide an introduction to the field of adaptive social robotics: \cite{ahmad2017systematic, martins2019user, nocentini2019survey}.
Adaptive behaviors are generally complex and difficult to engineer by hand, such as with finite state machines or behavior trees~\cite{colledanchise2018behavior}.

As a possible solution, RL~\cite{sutton2018reinforcement} can be used to learn complex social behaviors for robots from direct interactions with the environment.
But RL introduces its own challenges as discussed in the survey by Akalin and Loutfi~\cite{akalin2021reinforcement} which provides a general overview of the usage of RL in social robotics.
One crucial factor for the success of RL is the design of the reward function.
The reward function enables the optimal behavior to be learned by giving positive rewards for reaching desired states and behaviors and negative rewards for undesired ones.
Given a reward function, RL algorithms attempt to learn the optimal behavior by maximizing the reward.

As the behavior of an RL agent is determined by the reward function, adaptability can be introduced through two approaches:
Either 1) by including rewards for adaptability in the reward function, or 2) by adapting the reward function to each social situation.
A prominent line of research for the first direction is to obtain rewards directly from a human instructor who teaches the robot.
See for example \cite{patompak2020learning}, where human feedback is used to learn the optimal proxemics, i.e.,\ how close a robot should position itself to people, for a human-aware navigation controller. 
While collecting rewards from the user might be suitable for simple tasks requiring few examples to properly learn the task, it is not well suited to more complex learning problems requiring hundreds of samples.

Our work follows the second direction, where social environments and scenarios are modeled by different reward functions.
The goal of the RL agent is to adapt quickly to a new reward function.
An example of this method is the multitask learning approach in \cite{choi2020fast}, where reward functions are parameterized.
The agent learns a task-dependent policy, i.e.,\ a behavior that is conditioned to the parameters of the target reward function.
The policy is trained over several reward functions.
Given a new task and its reward function parameters, the agent is then directly adapted to it.
Nonetheless, this approach is restricted to parameterized reward functions, which are not always available. 
In this paper, we explore meta-RL methods for adaptive social robotics tasks, which have the advantage of adapting quickly to complex reward functions without the requirement of being parameterized.

\subsection{Deep Meta Reinforcement Learning}

Deep meta-RL aims to ``learn to reinforcement learn,'' which means to improve the learning speed in a new target task using experience from a set of similar meta-training tasks. 
The survey by Beck et al.~\cite{beck2023survey} categorizes algorithms by their goal into many-shot and few-shot algorithms.

Many-shot algorithms such as LPG~\cite{oh2020discovering}, or MetaGenRL \cite{kirsch2019improving}, try to improve standard RL algorithms by using many observations from the target tasks.
This stands in contrast with our goal of adapting quickly to a new social environment.

Few-shot algorithms such as MAML~\cite{finn2017model} or RL2~\cite{duan2016rl} aim to adapt to a new target task using only a few observations from that task.
MAML learns initialization parameters for a deep network which can then be trained faster in a target task.
MAML improves the learning performance, but it uses gradient-based updates to adapt to the target task, still requiring many observations and training iterations for adaptation.
In contrast, RL2 uses a recurrent deep network that is trained to identify the task characteristics from observations when executed in its target task, and then to adapt its own behavior to it.
RL2 does not require gradient-based updates to adapt and shows improved performance over MAML~\cite{rakelly2019efficient}.
A drawback of current MAML and RL2 implementations is that they are on-policy.
On-policy methods try to improve the policy that is currently used to collect environment observations \cite{sutton2018reinforcement}.
Conversely, off-policy algorithms such as Q-learning can optimize any policy (usually the optimal one) based on the collected observations.
This usually makes off-policy methods more sample-efficient than on-policy methods.

For this study, we choose to focus on PEARL~\cite{zhao2020meld, rakelly2019efficient}, an off-policy few-shot meta-RL algorithm.
Similarly to RL2, it uses observations from the target task to adapt directly to it without gradient-based updates. 
Given a target task, PEARL conditions its deep neural model to it with a description of the task in the form of a vector input $z \in \R^{d}$.
It utilizes a variational inference method to learn how to represent tasks and to identify a good task representation based on a few task observations. 
PEARL has shown that it needs 20 to 100$\times$ fewer observations to learn target tasks compared to MAML and RL2~\cite{rakelly2019efficient}, justifying our choice to use it as our base approach for meta-RL in social robotics.
We aim to evaluate the use and limitations of PEARL in meta-RL for social robotics and to provide technical solutions to address some of these limitations.

\subsection{Posterior Collapse}
As for any variational method, PEARL can suffer from posterior collapse, meaning that at least one of the latent bottleneck dimensions becomes non-informative during training.
This behavior can be easily identified by looking at the per-dimension Kullback-Leibler divergence between the posterior and prior distributions. 
Once a dimension collapses, it is not needed to ensure good reconstruction. 
Several factors can cause posterior collapse, e.g., a local optima~\cite{dai}, the objective function itself~\cite{zhao,razavi,sonderby,Huang}, a too unconstrained variance~\cite{reinke} or the fact that the posterior approximation lags behind the true posterior model~\cite{He}. 
In the case of PEARL, having collapsed dimensions translate into a reduction in the task identification power of the method. 
To mitigate the negative impact of posterior collapse, we propose to raise the representation power of the latent representation using a radial basis function layer.
In our experiments, this proves more beneficial than increasing the network capacity of PEARL.

\section{Approach}

\begin{figure*}
	\centering
	\includegraphics[width=1.0\textwidth]{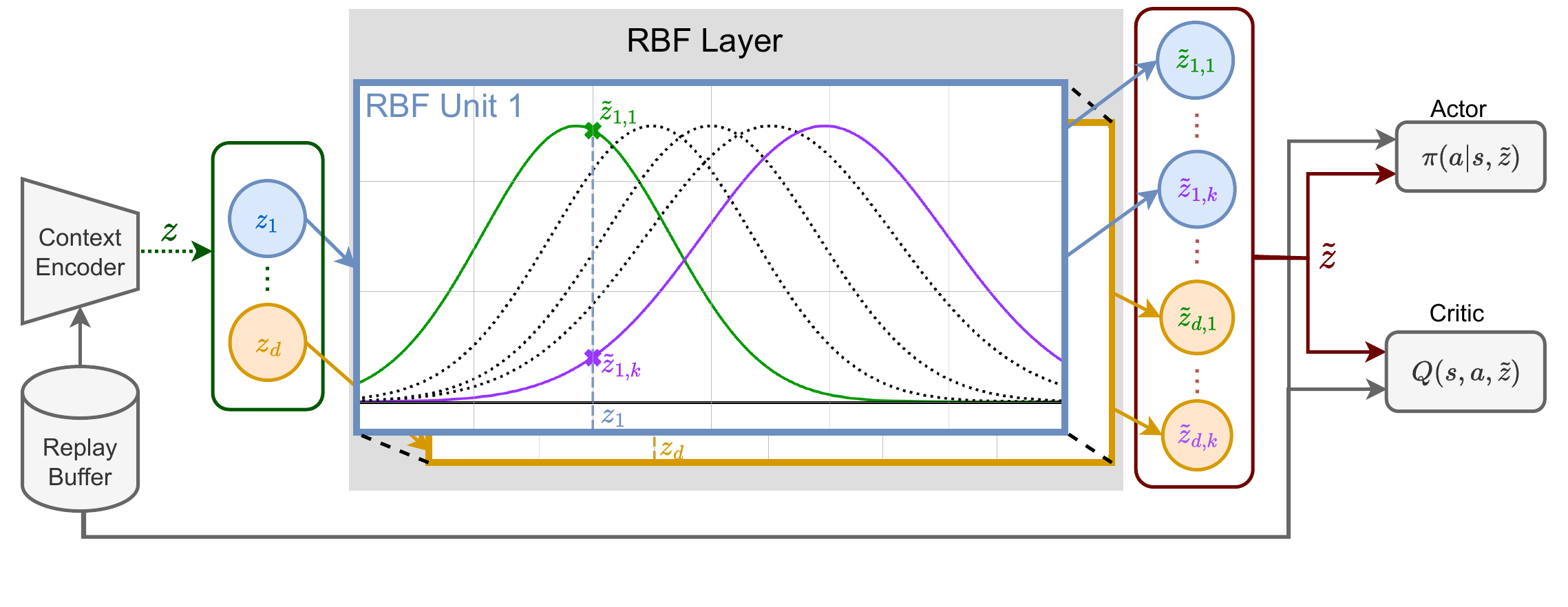}
	\caption{\textbf{PEARL-RBF Architecture.} The proposed meta-RL procedure learns to adapt to unseen test reward functions. The context encoder uses data from the replay buffer to infer the posterior over the latent context variable $z$. The latent context sampled from the posterior $z = (z_1,\ldots z_d)$ is then fed to the RBF network, which uplifts every input dimension $m$ to a different $k$-dimensional representation: $z_m\rightarrow (\tilde{z}_{m,1},\ldots,\tilde{z}_{m,k})$. The resulting task representation $\tilde{z}$ is used to condition the actor and critic network.}
	\label{fig:graph}
\end{figure*}   
\subsection{Preliminaries}
\paragraph{Reinforcement Learning} 

Tasks in RL are formalized as Markov decision processes (MDPs).
An MDP is a tuple $(\mathcal{S}, \mathcal{A}, \mathcal{P}, \mathcal{R})$ with state space $\mathcal{S}$ and action space $\mathcal{A}$.
The agent transitions from a state $s_t\in\mathcal{S}$ by action $a_t\in\mathcal{A}$ to state $s_{t+1}$ at time step $t$. The transition probability density function $\mathcal{P}(s_{t+1}\vert s_t, a_t)$ defines the environment dynamics giving the probabilities for transitions. 
For each transition, the agent receives a reward defined by the reward function: $r_t = \mathcal{R}(s_t, a_t, s_{t+1}) \in \mathbb{R}$. 
The probability transition function and reward function are unknown to the agent.
The goal of the agent is to maximize the expected future return for each time step $t$:
$G_t = \E\left[\sum^{\infty}_{k=0} \gamma^{k} \mathcal{R}(s_{t+k}, a_{t+k}, s_{t+1+k}) \right]$
where the discount factor $\gamma \in [0,1)$ defines the importance of short-term rewards relative to long-term ones. 
RL agents maximize the return by learning a policy $\pi(a \vert s) = \Pr(A_t = a \vert S_t = s)$ that defines the probability of the agent taking action $a \in \mathcal{A}$ in state $s \in \mathcal{S}$.

In the case of robotics tasks, the state is, for example, represented by the input from the robot's visual (cameras) and audio (microphones) sensors. Actions are the motor commands sent to its actuators. Taking an action will have an impact on the environment and will create stochastic change in it. The goal of the task is given by the reward function, e.g.,\ the robot is rewarded if it reaches a certain target position.

\paragraph{Variational Meta-RL (PEARL)}

In this work, we would like an agent to adapt quickly to a new reward function using only a few observations. 
We formalize this problem as a meta-RL problem.
Each reward function represents a task $\tau \in \mathcal{T}$ with $\tau = \{\mathcal{R}(s_t, a_t, s_{t+1})\}$.
We assume a distribution $\rho(\tau)$ over the task. 
We differentiate two meta-learning phases:
1) meta-training and 2) meta-testing.
During meta-training, a number of training tasks are sampled according to $\rho(\tau)$. 
Based on these training tasks, a task conditioned policy $\pi(a \vert s,\tau)$ is learned. 
During meta-testing, a different set of test tasks is sampled from $\rho(\tau)$ and the adaptation of the learned policy to these tasks is measured.

We chose PEARL \cite{rakelly2019efficient} as our meta-RL algorithm.
It learns a task-conditioned policy $\pi(a \vert s,z_\tau)$ where $z_\tau \in Z = \mathbb{R}^{d}$ is a low-dimensional task representation.
The task representation $z_\tau$ conditions the policy towards maximizing the return of the reward function for task $\tau$.
The representation $z_\tau = f(c^\tau)$ is computed based on observed transitions from a task, called the context $c^\tau=\{c^\tau_n\}$, where $c^\tau_n = (s_n, a_n, r_n, s'_n)$.
PEARL uses a variational method, similar to variational autoencoders (VAEs) \cite{kingma2013auto}, to estimate the posterior distribution of the low-dimensional representation given the context $p(z \vert c)$.
It approximates the posterior with an inference network $q_\phi(z \vert c)$ parametrized by $\phi$.
The network is trained on a log-likelihood objective resulting in the following variational lower bound:
\begin{equation}
    \label{eq:variational_lower_bound}
    \mathbb{E}_{\tau}\left[ \mathbb{E}_{z \sim q_\phi(z \vert c^\tau)} \left[\mathcal{R}(\tau, z) + \beta D_{KL}(q_\phi(z \vert c^\tau)) \Vert p(z) )\right] \right]
\end{equation}
where $\mathcal{R}(\tau, z)$ is the return for task $\tau$ using the policy conditioned on $z$ and $p(z)$ is a standard Gaussian prior over $z$. 
The inference network is a product of independent Gaussian factors for each transition in $c^\tau$:  
\begin{equation}
    q_\phi(z \vert c^\tau) \propto \prod_{n=1}^N \mathcal{N}\left(f_\phi^\mu(c^\tau_n), f_\phi^\sigma(c^\tau_n)\right)
\end{equation}
where $f_\phi^\mu$ and $f_\phi^\sigma$ are represented by a neural network.

During meta-training, the policy $\pi(a \vert s,z)$ is learned using the soft actor-critic (SAC) \cite{haarnoja2018softbis} algorithm. 
SAC is off-policy, consisting of an actor $\pi_{\theta_\pi}(a \vert s,z)$ and a critic $Q_{\theta_Q}(s,a,z)$ network. 
PEARL jointly trains the inference, actor, and critic networks using the reparameterization trick, similar to VAEs \cite{kingma2013auto}.
During a meta-training step, the training procedure has two phases: 1) data collection and 2) network parameter updating.
In the data collection step, a replay buffer  $\mathcal{B^\tau}$ is filled with the transitions from $K$ trajectories for each training task $\tau$. Then, for each trajectory PEARL samples a task representation $z \sim q_\phi(z \vert c^\tau)$ to condition the policy where $c^\tau$ is sampled from the replay buffer $\mathcal{B^\tau}$.
During the second phase, the procedure updates the network parameters for each training task $\tau$. It first samples a batch of context $c^\tau$ from recently sampled transitions in the replay buffer $\mathcal{B^\tau}$.
Then, the task representation $z \sim q_\phi(z \vert c^\tau)$ is sampled from the posterior distribution of $z$ given the context $c^\tau$.
The critic and actor networks are updated using the task representation and independently sampled transitions from the whole replay buffer.
The loss of the critic is given by:
\begin{equation}
    \mathcal{L}_{critic} = \E_{\substack{(s,a,r,s')\sim \mathcal{B} \\ z \sim q_\phi(z \vert c)} } \left[ Q_{\theta}(s,a,z) - (r+\dot{V}(s', \dot{z}))  \right]^2
\end{equation}
where $\dot{V}$ is the value, i.e.,\ the maximum Q-value, of a target network, and $\dot{z}$ indicates that gradients are not being computed through it. 
A target network is necessary here because directly implementing Q learning with neural networks was shown to be unstable in many environments~\cite{mnih2013playing}.
The actor loss is given by:
\begin{equation}
\mathcal{L}_{act} = \E_{\substack{s\sim \mathcal{B}, a\sim \pi_\theta \\ z \sim q_\phi(z \vert c)}} \left[ log(\pi_\theta(a \vert s, \dot{z}) - (Q_\theta(s, a, \dot{z})) \right]    
\end{equation}
The loss of the inference network for the task representation is composed of the critic loss and the  Kullback–Leibler divergence term from (\ref{eq:variational_lower_bound}):
\begin{equation}
    \mathcal{L}_{\phi} = \E_{\substack{(s,a,r,s')\sim \mathcal{B} \\ z \sim q_\phi(z \vert c)} } \left[   \mathcal{L}_{critic} + \beta D_{KL}(q_\phi(z \vert c^\tau)) \Vert p(z) )\right].
\end{equation}

For meta-testing, a test task $\tau$ is first explored for a few hundred time steps.
The policy $\pi_e(a \vert s,z_e)$ used for exploration is conditioned to a task representation sampled from the Gaussian prior $z_e\sim p(z)$.
After the exploration phase, the context $c^\tau$ collected with $\pi_e$ is used to compute the final task representation given by the sample mean of the posterior mean: $z^\tau = \frac{1}{N} \sum_{n=1}^N f^\mu_\phi(c^\tau_n)$. More details on the meta-testing phase are provided in the experimental section.

\subsection{RBF for Variational Meta-RL} 

We noticed that in several scenarios PEARL suffers from posterior collapse (of the learned task representation $z$).
As a result, not all dimensions of $z$ are used. The task information is compressed in a few dimensions, making it difficult for the downstream policy and critic network to learn from $z$. To compensate for this, we propose to transform the task representation $\tilde{z} = \varphi(z)$ to a representation that is easier to process for downstream networks. Inspired by~\cite{asadi2020deep}, we propose the usage of radial basis function (RBF) layers to transform the task representation (Fig.~\ref{fig:graph}).

We construct the RBF layer based on the idea of RBF networks \cite{broomhead1988radial}, which are universal function approximators \cite{park1991universal}.
RBF networks consist of a layer of hidden neurons that have a Gaussian activation function.
The output of the networks is a weighted sum over the Gaussian activations.
In a similar manner, our proposed RBF layer consists of a layer of neurons.
For each input dimension $z_{j} \in \mathbb{R} $ from the original task representation there exist $N$ RBF neurons: 
$\tilde{z}_{j,1}, \ldots, \tilde{z}_{j,N}$.
Each neuron represents a radial basis function having a Gaussian shape:
\begin{equation}
    \tilde{z}_{i,j} = \exp\left(-\delta_{i,j}  \vert \vert z_i-c_{i,j} \vert  \vert ^2\right),
\label{rbf-eq}
\end{equation}
where $\delta_{i,j} \in \mathbb{R}$ is a scaling factor and $c_{i,j} \in \mathbb{R}$ is the center, i.e.,\ the point of the highest activation.
$\delta$ and $c$ are the parameters of the RBF layer, which can be either fixed or trained using gradient descent based on the loss function of downstream networks. 

In summary, we propose to transform the task representation using an RBF layer: $\tilde{z} = \varphi(z)$.
The resulting representation is then given to the task conditioned actor $\pi_\theta(a \vert s,\tilde{z})$ and critic $Q(s,a,\tilde{z})$ network.

\section{Experimental Results}
\label{sec:result}

We evaluated our approach in three different environments inspired by social interaction scenarios. 
In the first environment, the agent, represented by a robotic head, needs to learn how to control its position to optimize the number of people in its field of view. 
In the second environment, the agent must learn to navigate safely and in a socially compliant manner through a crowd of people. 
The third environment is a continuous control environment focused on robotic locomotion. 
The purpose of these three environments is twofold. 
First, we want to show the effectiveness of our proposed methodology to generate different behaviors in human-robot interaction tasks. Second, we assess whether or not the use of RBF layers is beneficial to variational meta-RL in terms of training and adaptation efficiency.

\subsection{Evaluation protocol}
\paragraph{Baselines} 
Standard PEARL is a natural baseline to the proposed RBF-PEARL. 
However, directly comparing with PEARL seems unfair, since adding RBF layers increases the number of parameters of the actor and critic networks. 
We therefore adjust the number of parameters of the actor and critic networks of the PEARL baseline to match the number of parameters of the RBF-PEARL. 
Beyond this adjustment, both methods use separate actor and critic networks consisting of a 3-layer MLP with 300 neurons per layer. 
Both methods use a 3-layer MLP with 200 neurons per layer as the task encoder.
In more detail, if the latent dimension is $d$, and we use one RBF layer with $k$ neurons, the first layer of the actor/critic networks would have $300d$ parameters for PEARL and $300kd$ for RBF-PEARL. 
For a fair comparison, we add an extra layer at the beginning of PEARL's actor/critic network with $kd$ output neurons, rendering the number of PEARL's and RBF-PEARL's parameters comparable. 
Unless otherwise stated, we use $k=9$.

As a separate baseline, we used a modified version of the soft-actor-critic algorithm. 
We trained one agent per task, using only 200 observations. 
However, we strongly increased the number of gradient steps performed per observation in order to force the network to learn a behavior with only these 200 observations. 
We used this baseline to compare the performances of classical RL algorithms (here SAC) trained on 200 observations with respect to RBF-PEARL and PEARL with 200 adaption steps. 
We also compared the above algorithms against two state-of-the-art on-policy meta-RL methods: 
\begin{itemize}
    \item MAML-PPO, an on-policy gradient-based meta-RL algorithm that embeds policy gradient steps into the meta-optimization, and is trained with PPO ~\cite{finn2017model},
    \item RL2, an on-policy meta-RL algorithm that corresponds to training a GRU network with hidden states maintained across episodes within a task, trained with PPO~\cite{duan2016rl}.
\end{itemize}
For these two baselines, we used the implementation provided by the Garage reinforcement learning library ~\cite{garage}, a toolkit used to evaluate meta-RL and RL methods, providing state-of-the-art implementations. Both of these baselines are evaluated on the same number of environment steps as PEARL.

\paragraph{Procedure}	

All three environments are evaluated with the same procedure.
For meta-training, we sample 100 tasks. 
Evaluation is performed on 20 meta-testing tasks that are different from those used during meta-training. 
For PEARL and RBF-PEARL we collect observations for 200 time steps with a task representation $z_e$ sampled from the standard Gaussian prior on each meta-testing task $\tau$. 
The 200 observations are then used to compute the task representation $z^\tau$ sampled from the posterior estimation given by the encoder. 
To compute the final test-time performance, we record the performance of the policy associated with the task representation $z^\tau$ for one episode. 
We repeat the training process 5 times and report the mean performance and the associated standard error.

\subsection{Gaze control environment}
\label{c:gaze_control_env}

\begin{figure}[t]
  \centering
  \includegraphics[width=\columnwidth]{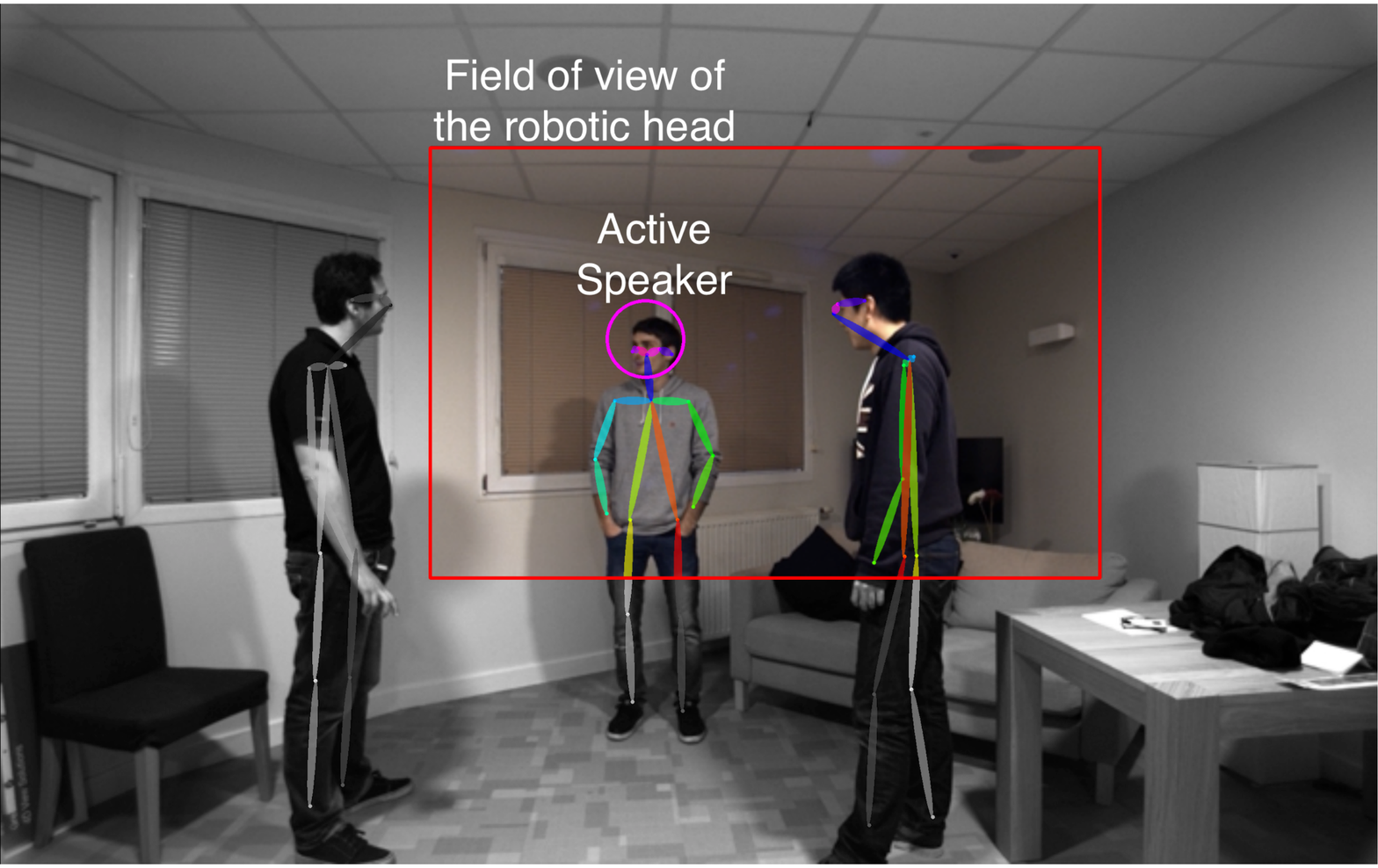}
  \caption{\textbf{Illustration of the gaze control environment.} The FOV is shown in red, the active speaker in purple, and the visible/non-visible landmarks are shown in white/color respectively. }
  \label{fig:gaze_env}
\end{figure}

\paragraph{Environment description} 
The goal in the gaze control environment is to learn a gaze strategy for a robotic head.
The environment is based on the work by Lathuiliere et al.~\cite{lathuiliere2019neural}.
The robot's observations are multimodal, consisting of visual, auditory, and proprioception cues. 
The entire visual scene has a size of $2\times 1$. 
The agent observes the scene with a head camera that extracts pose cues from a field of view (FOV) of the size $0.4\times 0.3$. 
The pose cues are represented by a heatmap for each of $J=18$ landmarks such as for the nose, neck, left shoulder, or right hip of people. 
The heatmap associated with each landmark indicates the probability of that landmark being present at every position. 
Since state-of-the-art pose estimators provide these heatmaps in low resolution, the visual observations will consist of $J$ $7\times 7$ heatmaps. 
Regarding the auditory observations, we emulate the output of a sound source localization algorithm. 
Given the precision of current sound source localization methods, it seems reasonable to represent the auditory input as a $14\times 8$ heatmap corresponding to the entire scene. 
Each of its cells corresponds to the probability of having an active speaker in this direction. 
Importantly, while the visual features correspond to the current FOV of the camera, the auditory features correspond to the entire scene, since audio localization is not limited by the camera's FOV.  
Lastly, the proprioception cues consist of the robot's head orientation, i.e.,\ the coordinates of the robot's current view center. 
It is encoded using a $\mathbb{R}^{2}$ vector representing both the pan and tilt angles of the robotic head.
An observation of the gaze control environment is shown in Figure~\ref{fig:gaze_env} (left).

The environment has a two-dimensional action space $[-1, 1]^2$, corresponding to the pan and tilt angular velocities. 
We choose to normalize the action space to stabilize the network learning. 
The maximum pan and tilt velocities correspond to shifting the camera FOV by $0.16$ and $0.11$ in scene coordinates at every time step.

\begin{figure*}[t] 
    \centering
    \includegraphics[width=\columnwidth]{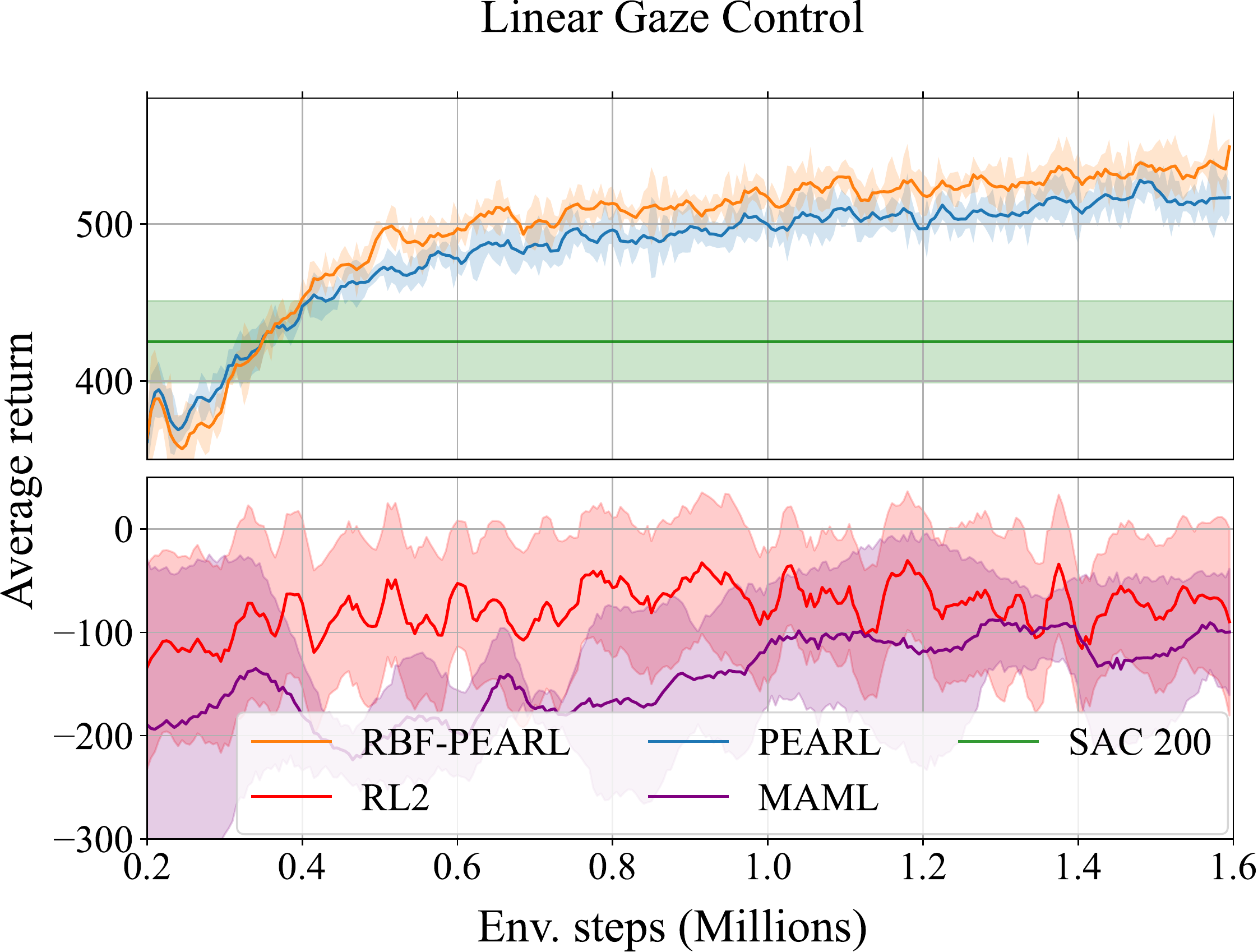}\hspace{3mm}
    \includegraphics[width=\columnwidth]{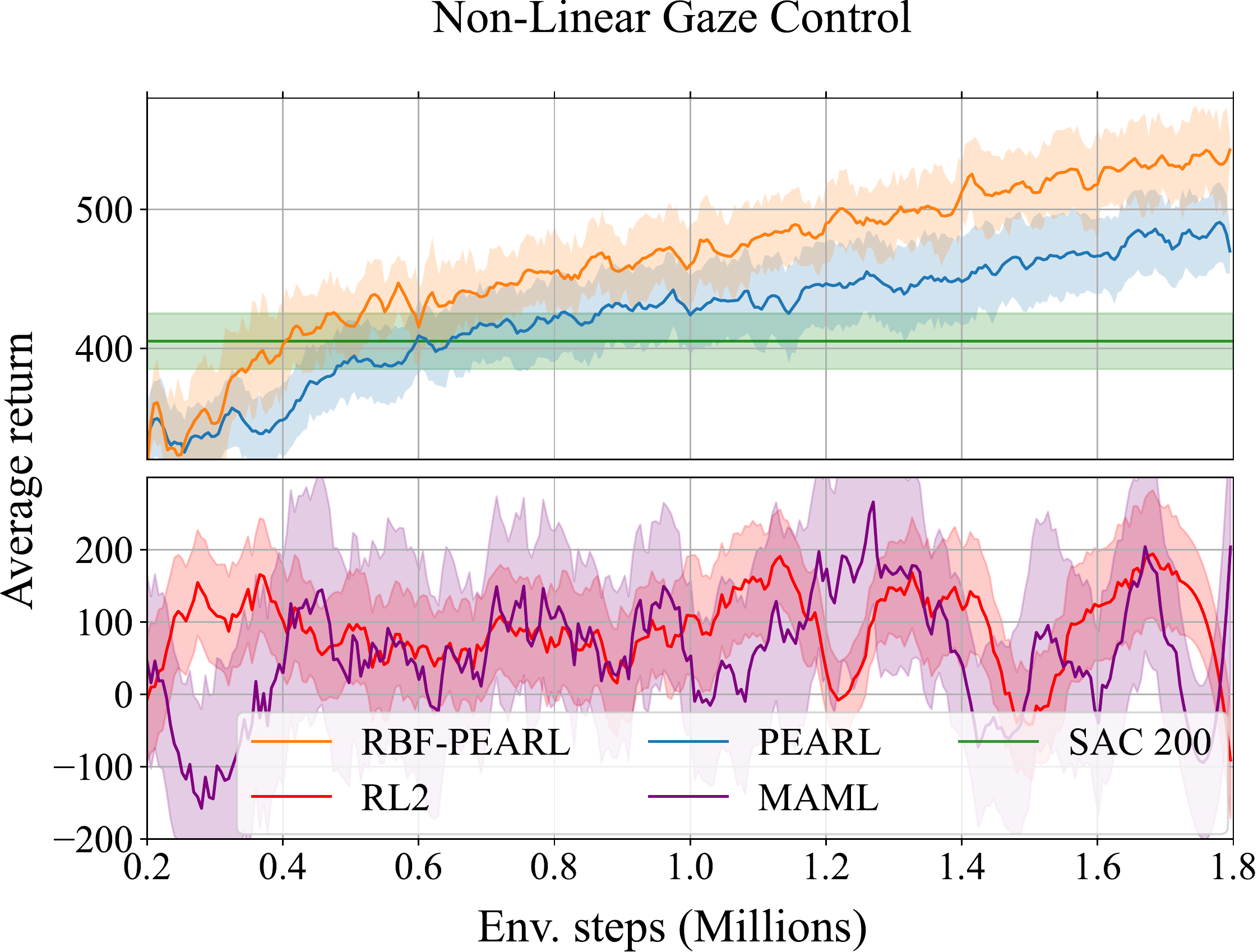} 
    \caption{\textbf{Results for the gaze control environment: RBF-PEARL outperforms PEARL by 30 and 70 in terms of average test-task return.} We plot the test-task performance (average return) vs.\ the number of collected samples during meta-training on the linear gaze control (left) and the non-linear gaze control (right) environments. In both cases, the off-policy meta-RL methods outperform SAC 200, which in turn outperforms on-policy meta-RL methods. 
    }
    \label{fig:gaze_results}
\end{figure*}

\paragraph{Reward components} 
We define three reward components that will be combined to generate various tasks (i.e.,\ reward functions). 
We will focus on the number of people in the FOV, the presence of a speaker in the FOV, and reducing spurious robot movements. 
The dimension of the latent space of both PEARL and RBF-PEARL is set to $d=3$.

We want to reward the robot for having people in the FOV of the camera since this means that the robot would be looking at people. 
Naively rewarding the number of people within the FOV leads to an action policy that explores until it finds a person, and then follows this person. 
A more natural behavior is to check on previously detected people. 
We therefore propose to use a visual reward component that depends on the last time the agent saw a person. 
More precisely:
\begin{equation}	
R_{\vis} = \sum_{p \in P_{\vis}} 2 - \exp(-t_p) 
\end{equation}
where $P_{\vis}$ is the set of people whose face is in the visual FOV and $t_p \in [0, \infty]$ is the time since the person was last seen (before the current frame). When a person remains in the FOV, the reward is close to $1$. When a person not seen for a long time reappears in the FOV, the reward is close to $2$. In this way, the desired behavior is encouraged.

As well as maximizing the number of visible people, we would like the robot to preferentially look at the speaking person(s).
Therefore, we define the audio component of the reward function as:
\begin{equation}
R_{\aud} = 
\begin{cases}
0    & \text{Nobody speaks} \\
-0.5 & \text{Speakers are outside the FOV}\\
2 \lvert P_{\aud} \rvert & \text{Speakers within the FOV}\\
\end{cases}
\end{equation}
where $P_{\aud}$ is the set of speakers in the FOV.

Finally, we would like to penalize expansive and brusque movements, since they are quite unnatural in social interactions. 
We define a negative movement component of the reward function, defined as:
\begin{equation}
    R_{\mov} = - K_{\mov} \sqrt{a_{pan}^2 + a_{tilt}^2}
\end{equation}
where $a_{pan}$ and $a_{tilt}$ are the two components of the action space, and $K_{\mov}=16$ is a constant to put $R_{\mov}$ in a similar numeric range to $R_{vid}$ and $R_{\aud}$.

\paragraph{Tasks} 
In order to construct different tasks (reward functions), we propose to use the defined components in two different ways. 
First, with simple combinations as in~\cite{lathuiliere2019neural}, and then with more complex ones (i.e.,\ non-linear). 
The first family of reward functions is generated by sampling convex combinations of the three components defined above: 
\begin{equation}
R^\tau = \omega_{\vis}^\tau R_{\vis} + \omega_{\aud}^\tau R_{\aud} + \omega_{\mov}^\tau R_{\mov} ~,
\end{equation}
where $\omega_{\vis}^\tau, \omega_{\aud}^\tau, \omega_{\mov}^\tau \in [0,1]$ are random convex weights, meaning that $\omega_{\vis}^\tau + \omega_{\aud}^\tau + \omega_{\mov}^\tau = 1$. 

We also wanted to compare the algorithms in more complex environments. 
To that end, we design our second family of reward functions using random multi-layer-perceptron (MLP) networks. 
These input the value of the three reward components defined above. 
The MLPs have 1 to 3 layers with 4 to 6 neurons each.
They have either no activation function with probability $0.25$ or a sigmoid with probability $0.75$.
The number of layers, the neurons per layer, the activation, and the weights are sampled randomly. 
Formally, we write:
\begin{equation}
	R^\tau = f^\tau(R_{vis}, R_{aud}, R_{mov}; W^\tau)
\end{equation}
where $f^\tau$ represents the sampled MLP network with sampled connection weights $W^\tau$.

\paragraph{Results}
We report the average return over the set of meta-testing tasks over the meta-training iterations (Figure~\ref{fig:gaze_results}). 
More precisely, we plot the average return mean and standard deviation over the five independent runs, for the gaze control environment with convex (top) and non-linear (bottom) combinations of reward components.

Generally, both PEARL and RBF-PEARL are able to provide a better adaptation starting point with the training progress. 
In addition, we observe that RBF-PEARL has a steeper learning curve than PEARL on both types of reward functions. 
More precisely, for convex combinations of reward components, RBF-PEARL performs comparably to PEARL during the first $400$k steps. 
From this point on, RBF-PEARL systematically outperforms PEARL by a margin of 20-30. 
For the non-linear combinations of reward components, RBF-PEARL exhibits superior performance from $600$k meta-training steps on, by a margin of roughly 50. 
Overall, in these two families of tasks, RBF-PEARL is faster than PEARL and shows a better asymptotic performance, thus showcasing the benefit of using the RBF layer.
As expected, both PEARL and RBF-PEARL achieve better performances than the batch of agents trained using the soft actor-critic algorithm with only 200 observations (SAC 200). 
In the Linear Gaze Control environment, the average performance achieved by agents trained with the soft-actor-critic algorithm is inferior to the performances of PEARL and RBF-PEARL by a margin of 100.
In the non-linear environment, the gap is lower between PEARL and SAC 200 as PEARL outperforms SAC 200 by a margin of 50. 
With RBF-PEARL, the performance gap with SAC 200 is more significant, as RBF-PEARL exhibits a superior performance by a margin of 130.

We observe that PEARL significantly outperforms on-policy meta-RL baselines in both tasks, as shown by the final performances after 1.6 and 1.8 million environment steps. 
More precisely, in the linear environment, the average performance of MAML is -99, and the average performance of RL2 is -89. 
In the non-linear environment, the average performance of MAML is 203, and the average performance of RL2 is -90. 
These poor performances are most likely due to the poor sample efficiency of on-policy methods.

\subsection{Social navigation environment}
\paragraph{Environment description}

\begin{figure}[t] 
    \centering
    \includegraphics[width=.35\textwidth]{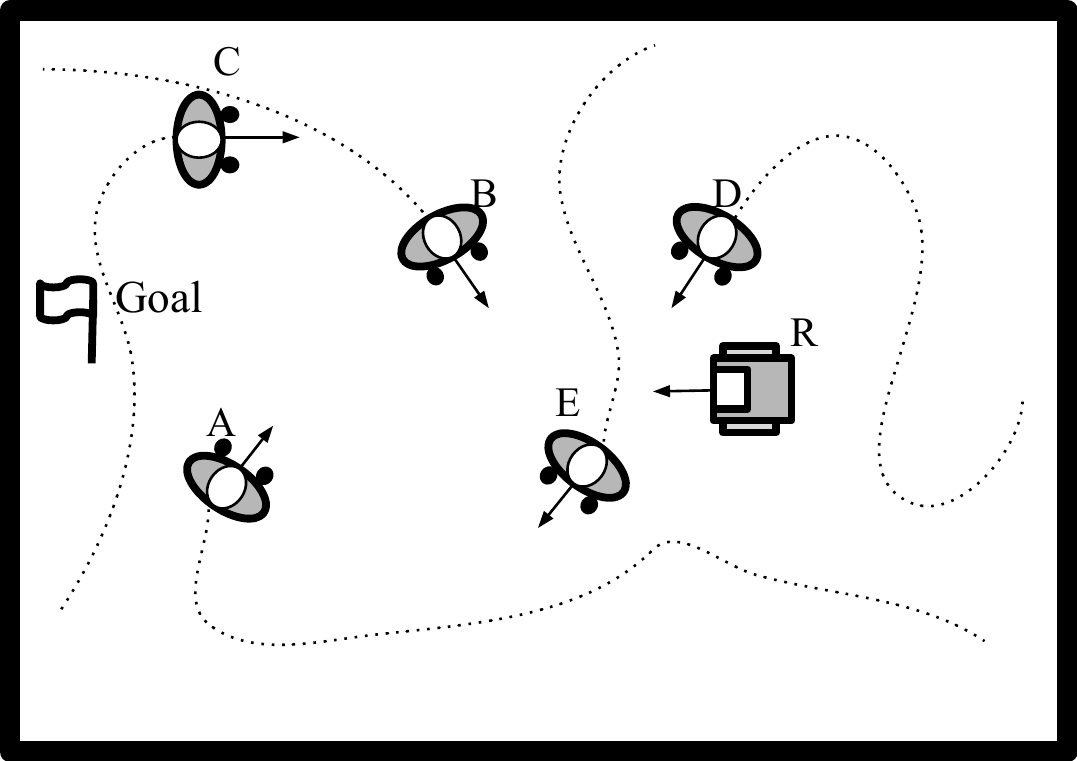} 
    \caption{
    \textbf{Schematic representation of the social navigation environment}. 
    The robot $R$ must move towards its goal position while navigating around five human agents ($A, \ldots, E$).
    }
    \label{fig:social_nav_env} 
\end{figure}

In this environment, the agent has to learn a human-aware navigation strategy~\cite{moller2021survey}. 
Our simulation environment is an empty room of dimension $15 \times 10$~m (Figure~\ref{fig:social_nav_env}). 
The room is populated with five human agents, whose position at the beginning of each episode is randomly initialized. 
Likewise, we randomly sample a robot target position at the beginning of an episode. 
The robot should reach the target position before the end of the episode without disturbing the human agents. 
Each human agent is given a random target position that they will go towards. 
If the robot reaches its goal it will be assigned a new one. 
To simulate the behavior of human agents, we model the human agents' motion using a social force model~\cite{pedica2008social} to generate plausible trajectories. 
This framework also limits the number of collision between human agents.

The robot agent has access to the coordinates of all people in the scene, their velocities, and their orientations. 
The robot also has access to its own velocity, coordinates, and target position. 
While the robot always begins the episode at the same position (coordinates $(14, 5)$), its target position is randomly sampled following $x \sim \mathcal{U}(0.0, 2.0)$ and $y \sim \mathcal{U}(0.0, 10.0)$, for the $x$ and $y$ coordinates, respectively. 

The action space is a continuous two-dimensional space set to $[-15, 15] \times [-2, 2]$, which corresponds respectively to the angular velocity in $rad.s^{-1}$ and linear velocity in $m.s^{-1}$. 
The maximum value of the linear and angular velocity is chosen such that the robot can go as fast as any human agent in the scene. 
In this environment, we use a smaller number of neurons $k = 5$ for our RBF layer.

\paragraph{Reward components}
We define a set of five reward components that will be combined to generate various tasks. 
The dimension of the latent space of both PEARL and RBF-PEARL is set to $d=5$ for this environment.

First, the goal component $R_g$ rewards the agent for reaching the target position:
\begin{equation}
	R_{g} = 1-\frac{d(r, g)}{D},
\end{equation} 
where $d(r,g)$ is the distance between the robot and the goal and $D$ is a normalizing factor to guarantee that the goal component stays within $[-1,1]$.
    
Second, the collision component $R_c$ penalizes collisions between the robot and human agents:
\begin{equation}
R_{c} = 
\begin{cases}
0    & \text{$d(r, h_i) > d_c$} \\
-1 & \text{$d(r, h_i) < d_c$}\\
\end{cases}
\end{equation}
where $d(r, h_i)$ is the distance between the robot and the human agent $i$ and $d_c$ is the collision threshold between the robot and the human agent.
        
Third, the social component $R_s$ rewards the robot for maintaining a safe distance away from all human agents. 
The social component depends on the distance between the robot and each human agent. 
If the distance between the robot and one of them is below a certain threshold, the robot will be penalized. 
The closer the robot is to a person, the higher the penalty will be. 
If the robot is close to more than one human agent, only the closest person to the robot is taken into consideration (that is, the human agent that will generate the lowest reward):
\begin{equation}
    R_{s} = \min\limits_{i} \left[\frac{d(r, h_i)}{d_{s}} - 1\right]
\end{equation}
where $d(r, h_i)$ is the distance between the robot and the human agent $i$ and $d_s$ can be understood as a threshold. 
Indeed, if the minimum distance between the robot and a human is below $d_s$, the reward becomes negative. 
Thus, $d_s$ can be seen as the distance at which the robot enters the comfort zone of people.

Fourth, the approach component $R_a$ is designed to reward the robot for positioning itself to avoid making other human agents aware of it. This component is based on the approach by Satake et al.~\cite{satake2012robot}, who quantify how aware people are of a robot when it approaches them.
The awareness is based on the relative positions and orientations of the robot and the human agents.
We compute the robot's awareness of each human agent and reward a low awareness of the robot. 
Awareness is a combination of visibility $R_{\text{visible}, i}$ and direction $R_{\text{direction}, i}$. 
While visibility assesses how visible the robot is to the human agent, direction assesses if the robot is going in a direction that would make the human agent more aware/afraid of it. 
    
Visibility for human agent $i$ is defined as:        
\begin{equation}
R_{\text{visible}, i}=
\begin{cases}
1-\frac{\theta_{i, r}}{\theta_{th}} & \theta_{i, r} < \theta_{th}\\
-\frac{\theta_{i, r}-\theta_{th}}{\pi - \theta_{th}} & \text{otherwise}
\end{cases}
\end{equation}
where $\theta_{th}$ is a threshold angle from which the robot is visible to the human agent and $\theta_{i, r}$ is the angle of the robot relative to the $i$'s human agent motion direction. 
If the robot is in front of the human agent, the value will be close to $1$, and if the robot is behind the human the value will be close to $-1$. 

The direction for human agent $i$ is defined as:
\begin{equation}
R_{\text{direction}, i} = 
\begin{cases}
1-\frac{\theta_{i, r}}{\pi/2} & \theta_{i, r} < \theta_{th}\\
1 & \text{otherwise}
\end{cases}
\end{equation}
It is designed to address how the human agent perceives the robot coming toward him. 
If the robot is coming closer to the human agent, he/she will become more aware of it and be distracted by it. 
On the other hand, even if the robot is visible to the human agent, if it moves away from him/her, the human agent will be less likely to be distracted by the robot.

To compute the approach reward component, we compute the minimum over the visibility and direction product over the human agents:
\begin{equation}
R_{a} = \min_{i} R_{\text{visible}, i} \cdot R_{\text{direction}, i}
\end{equation}

\begin{figure}[t]
    \centering
    \includegraphics[width=.48\textwidth]{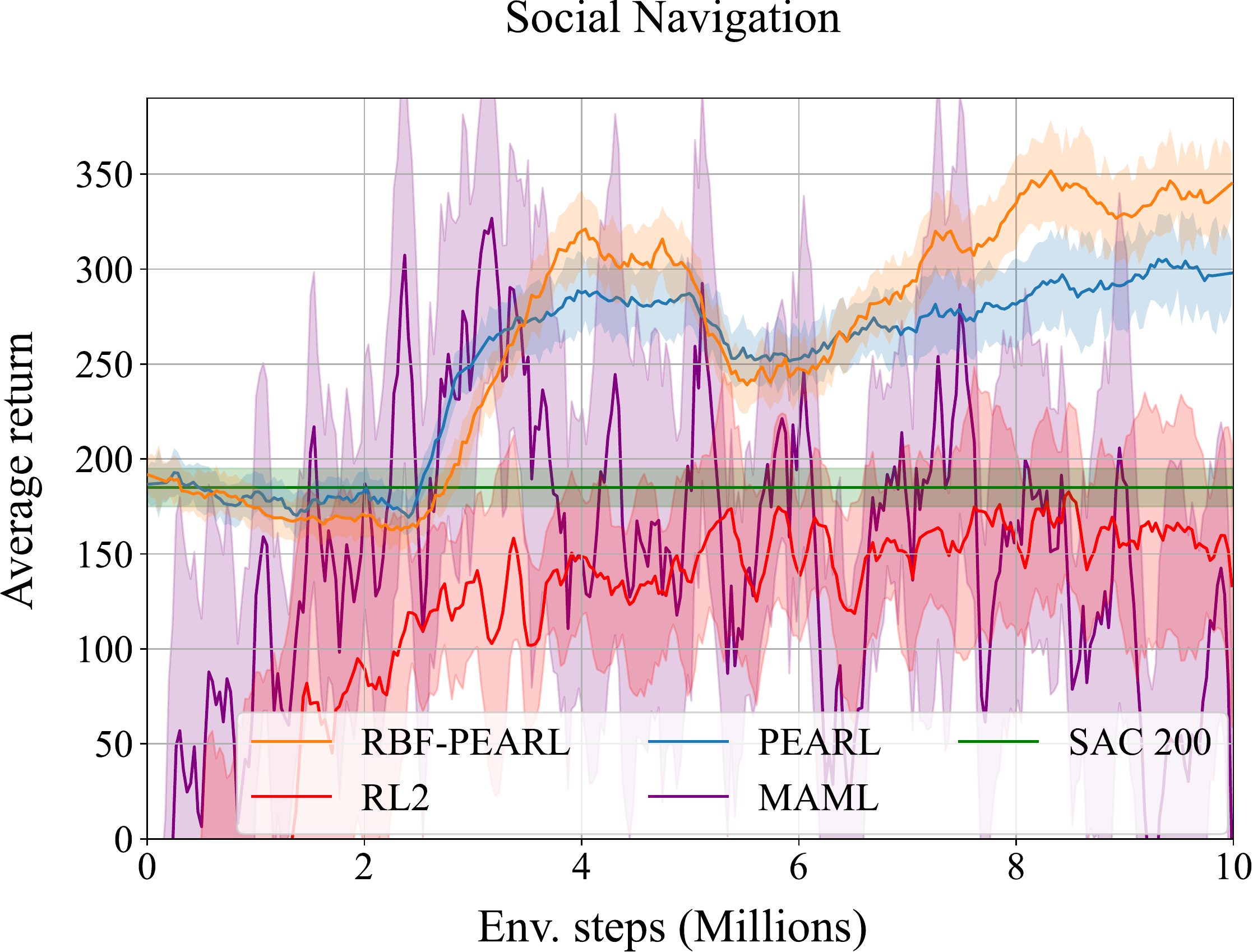}
    \caption{\textbf{Results on the social navigation environment: PEARL-RBF outperforms PEARL-RBF in average test-task performance by 40.} We plot the average test-task return vs.\ the number of samples collected during meta-training on the social navigation environment. Both off-policy meta-RL methods outperform SAC 200, which in turn outperforms both on-policy meta-RL methods.
    }
    \label{fig:social_nav_results}
\end{figure}

Fifth, and last, the velocity component $R_v$ is designed to penalize the robot for going too fast when it is in front of people. 
It is preferable to avoid having the robot moving fast in front of people as they may be disturbed or even afraid of it. 
The velocity component therefore penalizes the robot for moving fast if it is visible to human agents $i$:
\begin{equation}
R_{v, i }=
\begin{cases}
-e^{v}(1-\frac{\theta_{i, r}}{\theta_{th}}) & \theta_{i, r} < \theta_{th}\\
0 & \text{otherwise}
\end{cases}
\end{equation}
where $\theta_{th}$ is a threshold angle from which the robot is visible and $\theta_{i, r}$ is the angle of the robot relative to the human agent's motion direction. 
To compute the reward component, we adopt the same methodology as with the social and approach component.
We choose the minimum value obtained from all the human agents in the scene:
\begin{equation}
R_{v} = \min_{i} R_{v, i}.
\end{equation}

\paragraph{Tasks} 
We evaluate the meta-RL algorithms on reward functions that are convex combinations of the five reward components: 
\begin{equation}
	R^\tau = \omega_{g}^\tau R_{g} + \omega_{c}^\tau R_{c} + \omega_{s}^\tau R_{s} + \omega_{a}^\tau R_{a} + \omega_{v}^\tau R_{v}~,
\end{equation}
where $\omega_{g}^\tau, \omega_{c}^\tau, \omega_{s}^\tau, \omega_{a}^\tau, \omega_{v}^\tau \in [0,1]$ are the randomly sampled convex weights leading to task $\tau$, meaning that $\omega_{v}^\tau + \omega_{c}^\tau + \omega_{s}^\tau + \omega_{a}^\tau + \omega_{v}^\tau = 1$. 
The social navigation environment is more challenging than the gaze control environment.
The higher number of reward components makes the learning process more difficult, as well as behavior generation as there is a wider variety of possible reward functions.

\paragraph{Results}

We report the average return over the meta-testing tasks with the progress of meta-training for PEARL and RBF-PEARL (Figure~\ref{fig:social_nav_results}). 
As in the previous environment, we report mean and standard deviation over five runs. 
Similarly to the previous case, the performance of the two methods looks similar during the first steps of the training.
More importantly, the RBF layer seems to have a positive impact on the asymptotic performance. 
Indeed, after $6$ million steps, the RBF-PEARL algorithm performs better than PEARL.
Similarly to the gaze control environment, both PEARL and RBF-PEARL perform better than SAC-200 by a large margin (170).
They also outperform both on-policy methods, MAML and RL2, by a margin of 304 and 164, respectively.

\subsection{Racer environment}

\paragraph{Environment description} 

\begin{figure}[t] 
    \centering
    \includegraphics[width=.25\textwidth]{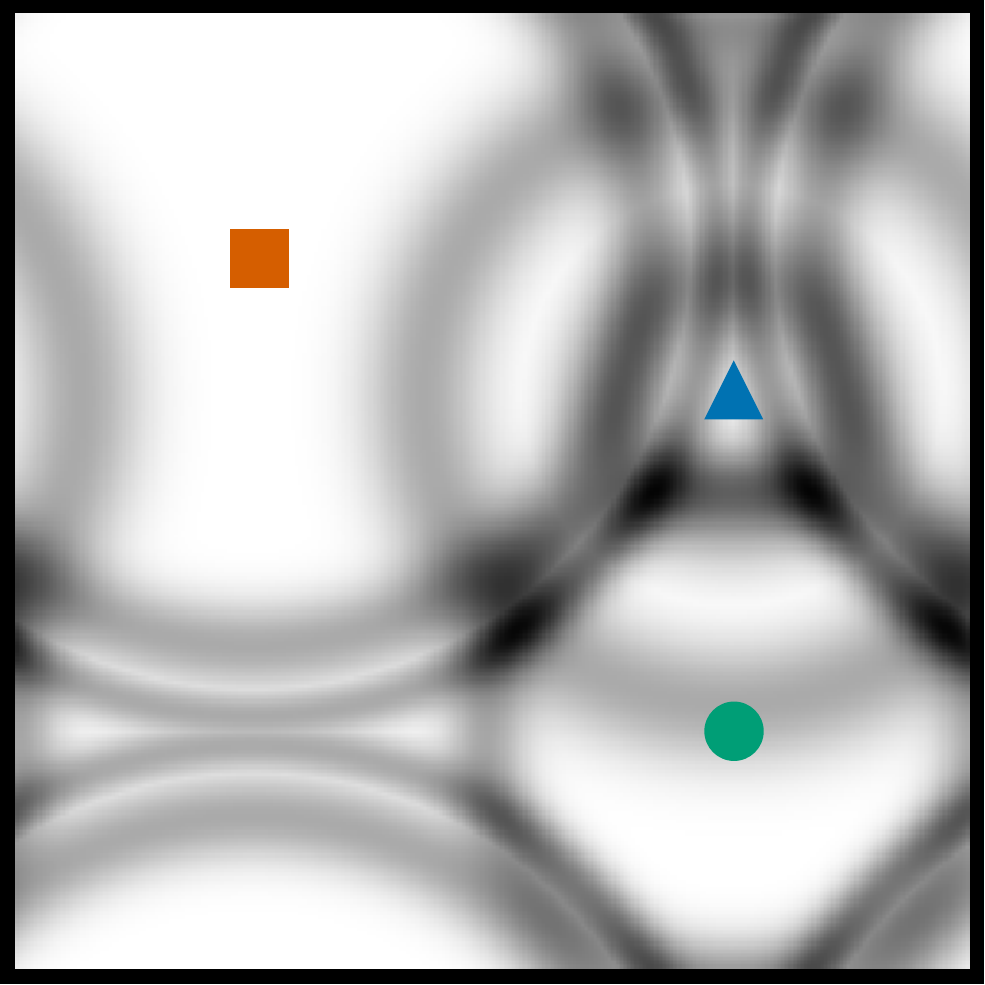} 
    \caption{\textbf{Illustration for the racer environment.} The three markers are depicted in blue, green, and orange. Dark regions correspond to high-reward regions. In the task depicted in the figure, the red and blue markers have two Gaussians, while the green one has only one. The number of Gaussians, their mean, and standard deviation are randomly sampled for each task.
    }  
    \label{fig:racer_env} 
\end{figure}

We evaluated the algorithms in an additional non-social task with complex, non-linear reward functions, called the racer environment \cite{reinke2021xi}. 
The agent has to navigate in a continuous two-dimensional scene for two hundred time steps (Fig.~\ref{fig:racer_env}). 
Similar to a car, the agent has an orientation and momentum, so that it can only drive straight, or around a right or left curve. 
The agent reappears on the opposite side if it exits one side. 
At the beginning of an episode, the agent is randomly placed in the environment. 
The agent's state is a vector $s \in \mathbb{R}^{120}$ corresponding to the agent's position and orientation. 
The position is encoded using a $10\times10$ evenly distributed grid of two-dimensional Gaussian radial basis functions. 
Similarly, the orientation is also encoded using $20$ Gaussian radial basis functions.
The action space of the agent consists of a one-dimensional continuous space set to $[-1, 1]$. 
The value of the action corresponds to the force applied to the agent, which then modifies the agent's orientation and position. 
For example, if the value of the action is close to -1, the agent will make a left curve. 

\paragraph{Reward components}
We define three reward components, each of them associated with one of the three markers in Figure~\ref{fig:racer_env}. 
More precisely, each reward component $r_k$ is defined as the maximum over Gaussian-shaped functions over the distance to the $k$-th marker $d_k$:
\begin{equation}
\label{eq:racer_env_reward_func}
    r_k = \max \left\{ \exp\left(-\frac{(d_k - \mu_{k,j})^2}{\sigma_{k,j}}\right)  \right\}_{j=1}^{n_k}
\end{equation}
where $n_k$ is the number of Gaussians for marker $k$, and $\mu_{k,j}$ and $\sigma_{k,j}$ are the mean and standard deviation of the $j$-th Gaussian of marker $k$. 
For each task, the parameters of the reward components are randomly sampled, as explained below.

\begin{figure}[t] 
    \centering
    \includegraphics[width=.45\textwidth]{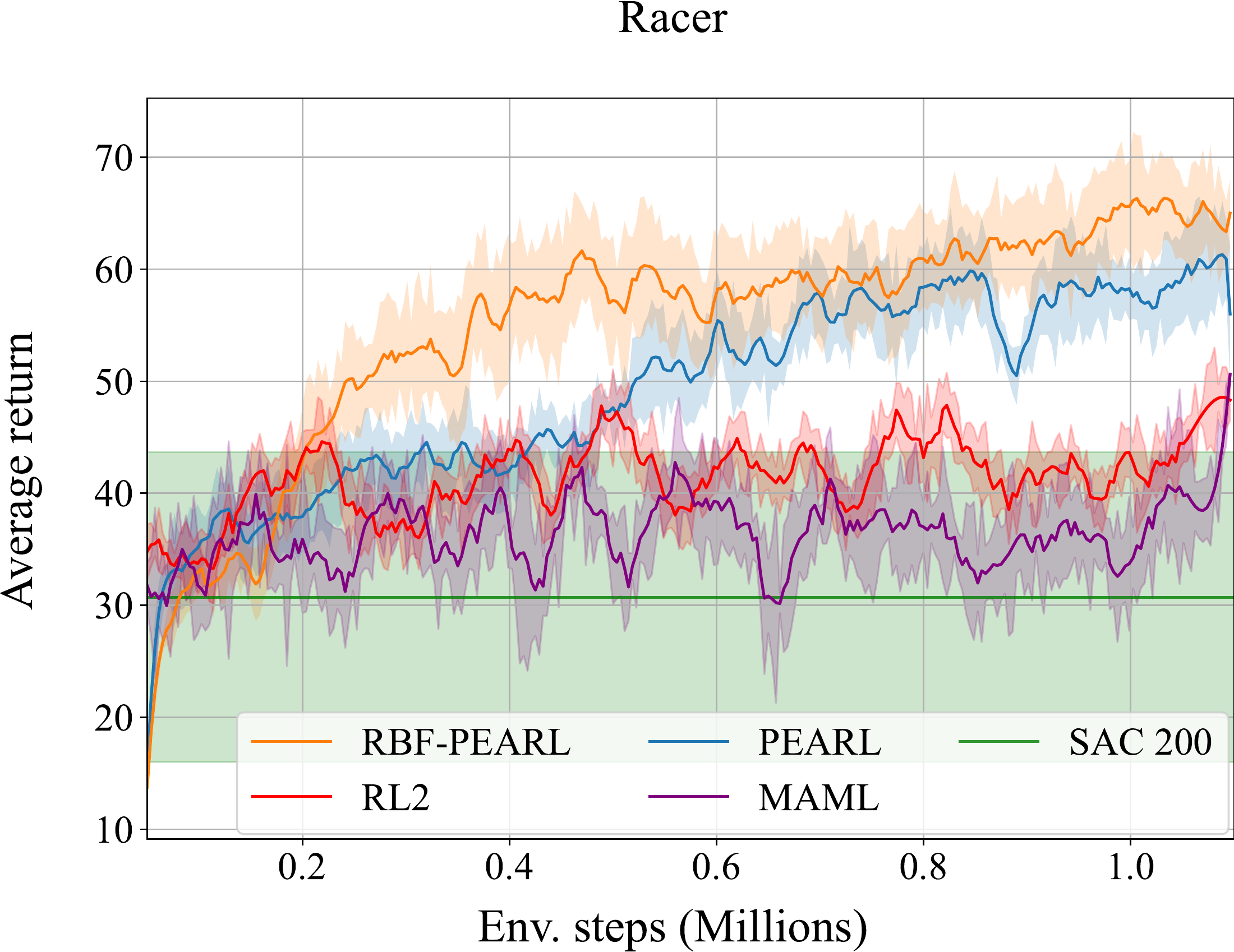} 
    \caption{\textbf{Results on the racer environment: PEARL-RBF outperforms PEARL in average test-task return by 10.} We plot the average test-task performance vs.\ the number of samples collected during meta-training. Off-policy meta-RL methods outperform on-policy meta-RL methods and SAC 200. It is the only environment where on-policy meta-RL outperforms the SAC 200 baseline.
    } 
    \label{fig:racer_results} 
\end{figure}
	
\paragraph{Tasks} 

The tasks differ in terms of the parameters of each of the three reward components. 
The number of Gaussians is sampled uniformly: $n_k \sim \mathcal{U}\{1,2\}$. 
The two parameters of each Gaussian component are sampled according to $\mu_{k,j} \sim \mathcal{U}(0.0, 0.7)$ and $\sigma_{k,j} \sim \mathcal{U}(0.001, 0.01)$. 
This sampling instantiates the three reward components for task $\tau$, $r_k^\tau$, and the final reward function is written:
\begin{equation}
    R^\tau= \frac{1}{3}\sum_{k=1}^3 r^\tau_k(d_k).
\end{equation}
The dimension of the latent space of both PEARL and RBF-PEARL is set to $d=3$.

\paragraph{Results}

RBF-PEARL consistently shows a stronger performance than PEARL in the racer environment (Fig.~\ref{fig:racer_results}).
After $300,000$ steps, RBF-PEARL constantly outperforms PEARL. 
The asymptotic performance of RBF-PEARL is also higher than PEARL. 
The average return at the end of the training is $54 \pm (11)$ for PEARL and $65 \pm (2)$ for RBF-PEARL. 
Also, both PEARL and RBF-PEARL (56 and 65 respectively) outperform SAC 200 by a large margin, performing two times better: SAC 200 obtains a performance of only 30.
Once again, we notice that PEARL outperforms on-policy meta-RL baselines.
The final performance of MAML and RL2 after 1.1 million environment steps is 51 and 48, respectively, far below the performance of PEARL for the same number of environment steps.

\section{Discussion}

\begin{figure*}[t]

\centering

\includegraphics[width=.32\textwidth]{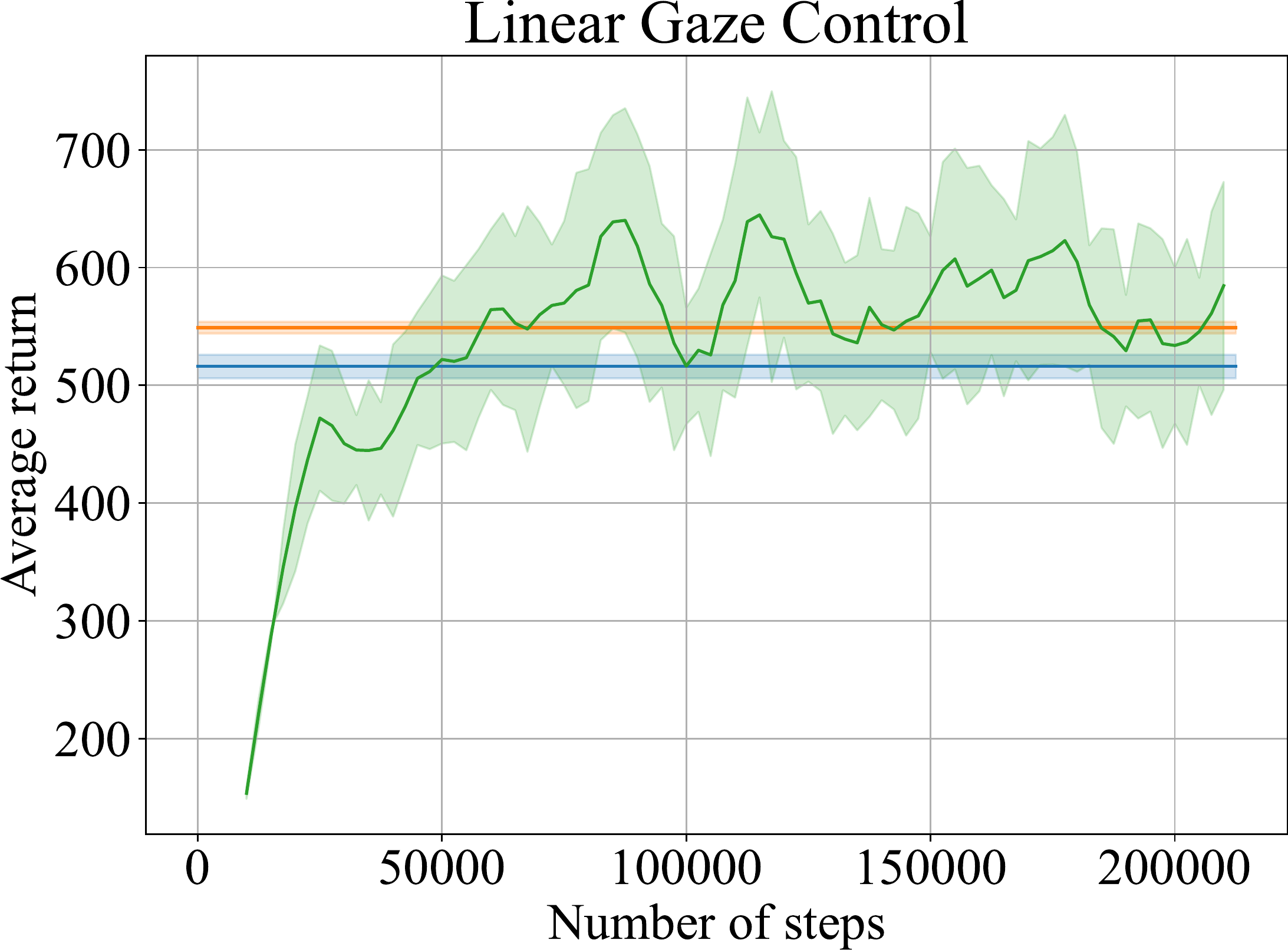}\hfill
\includegraphics[width=.32\textwidth]{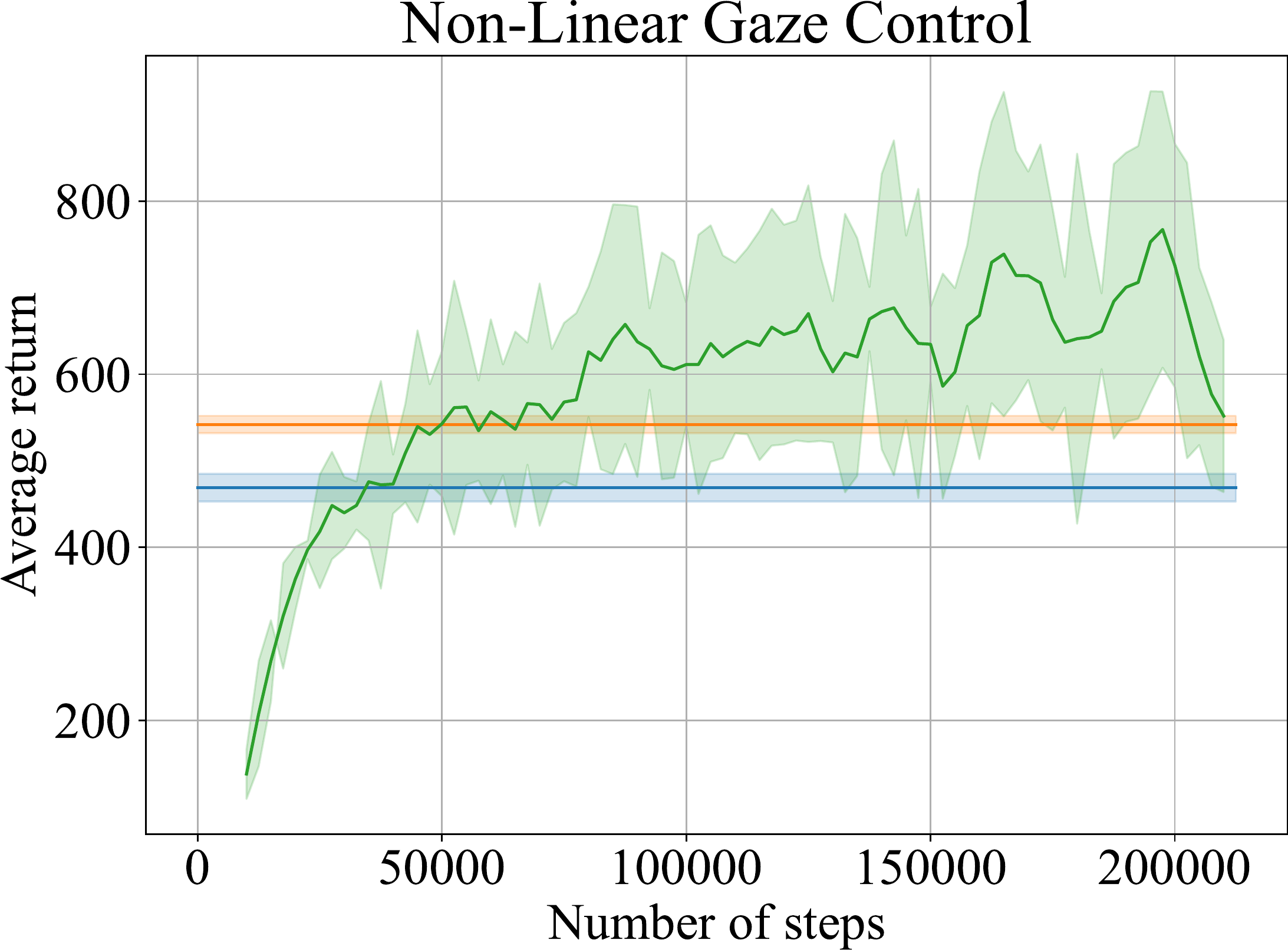}
\includegraphics[width=.32\textwidth]{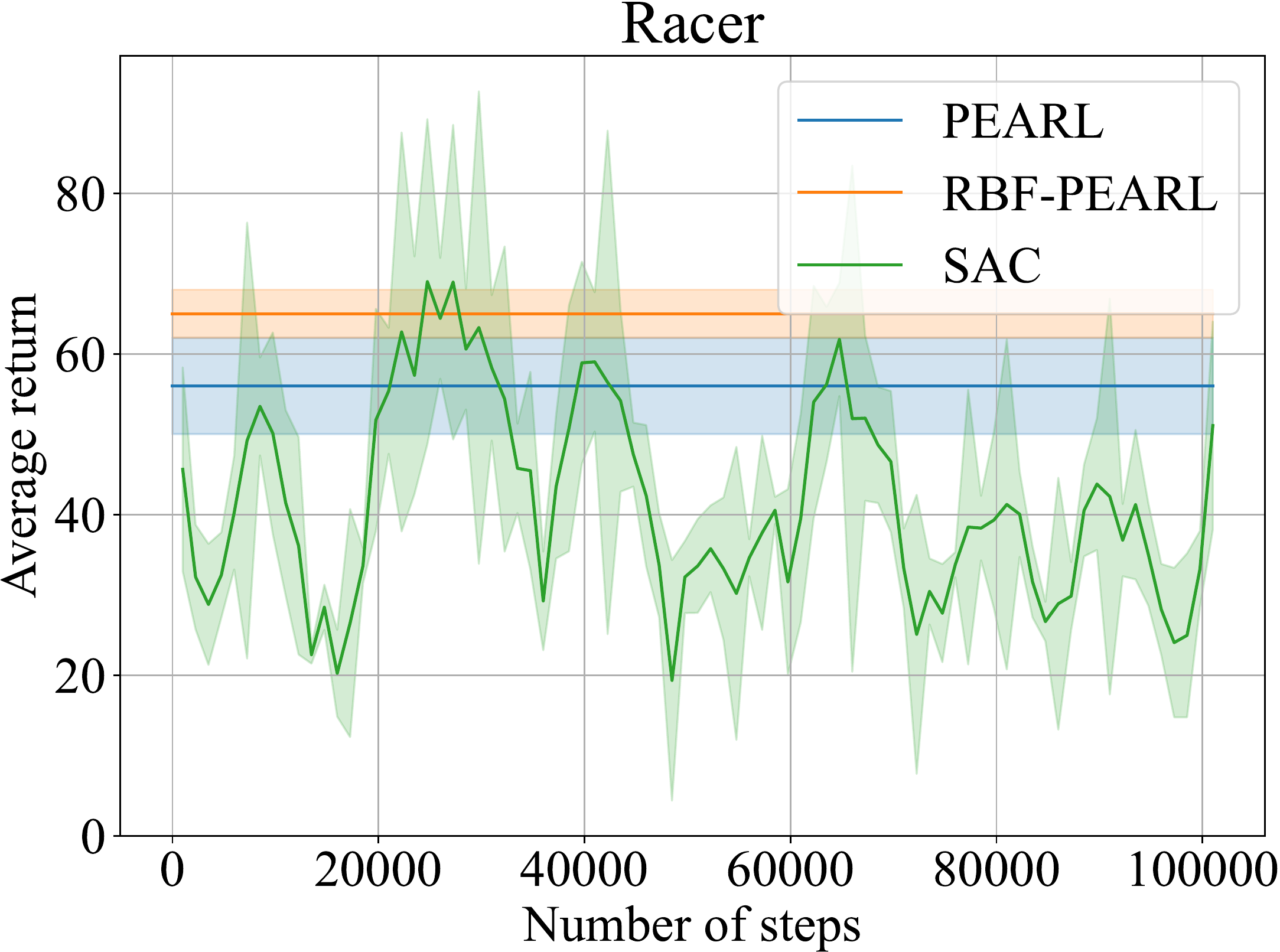}\hfill

    \caption{\textbf{Performances achieved by PEARL and RBF-PEARL compared to full training with SAC: SAC needs more than $100\times$ environment steps.}
    In order to reach the performance of PEARL/PEARL-RBF, SAC needs several thousand environment steps, compared to the 200 environment steps needed by PEARL/PEARL-RBF after the meta-training phase.  
    Results show the average test-task performance and standard deviation per environment step over 20 tasks and 5 seeds.  
    }
\label{fig:sac_vs_metarl}
\end{figure*}

\begin{figure}[t]
\begin{subfigure}{.475\linewidth}
  \includegraphics[width=\linewidth]{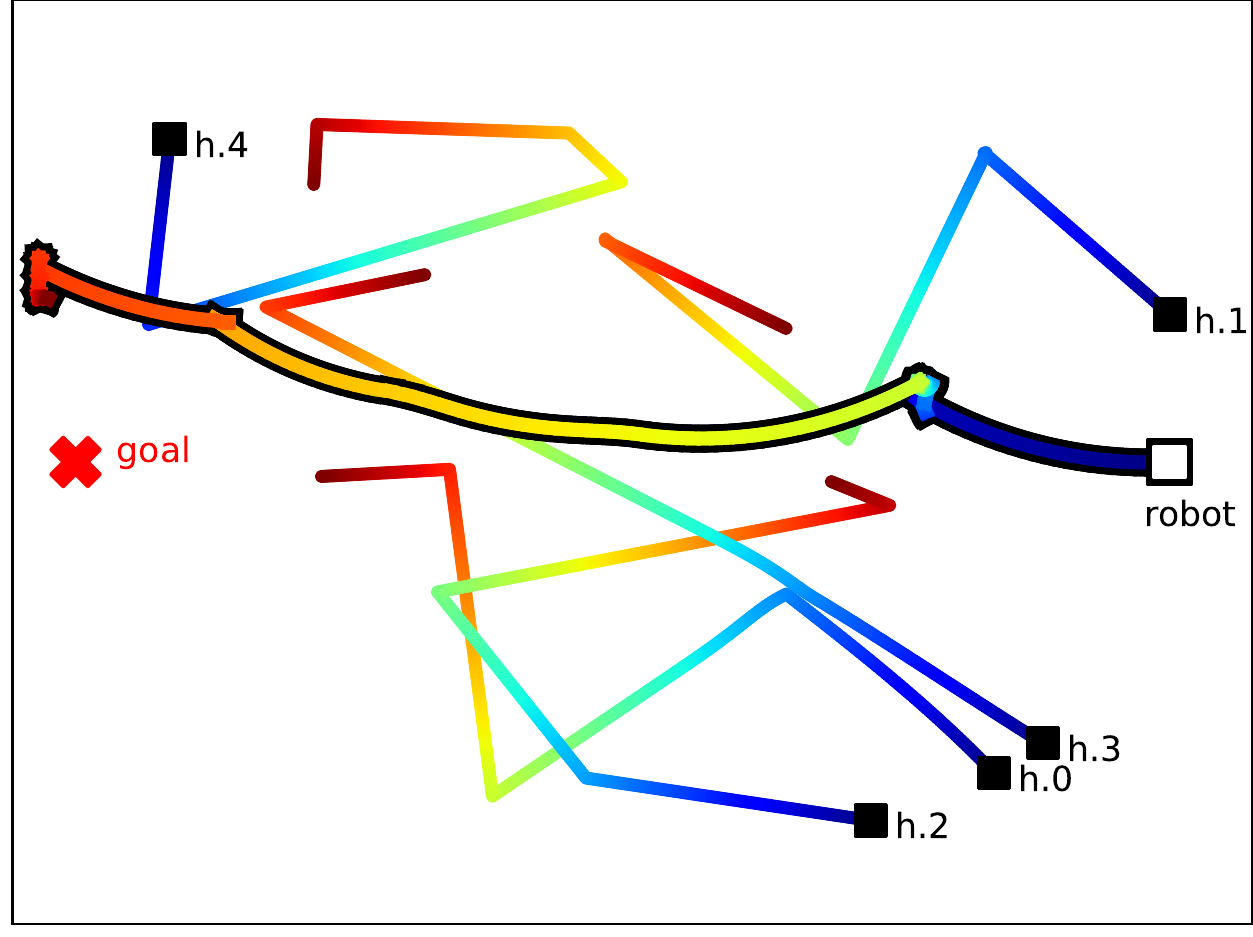}
  \caption{weights \& goal: 1.0 }
  \label{MLEDdet}
\end{subfigure}\hfill 
\begin{subfigure}{.475\linewidth}
  \includegraphics[width=\linewidth]{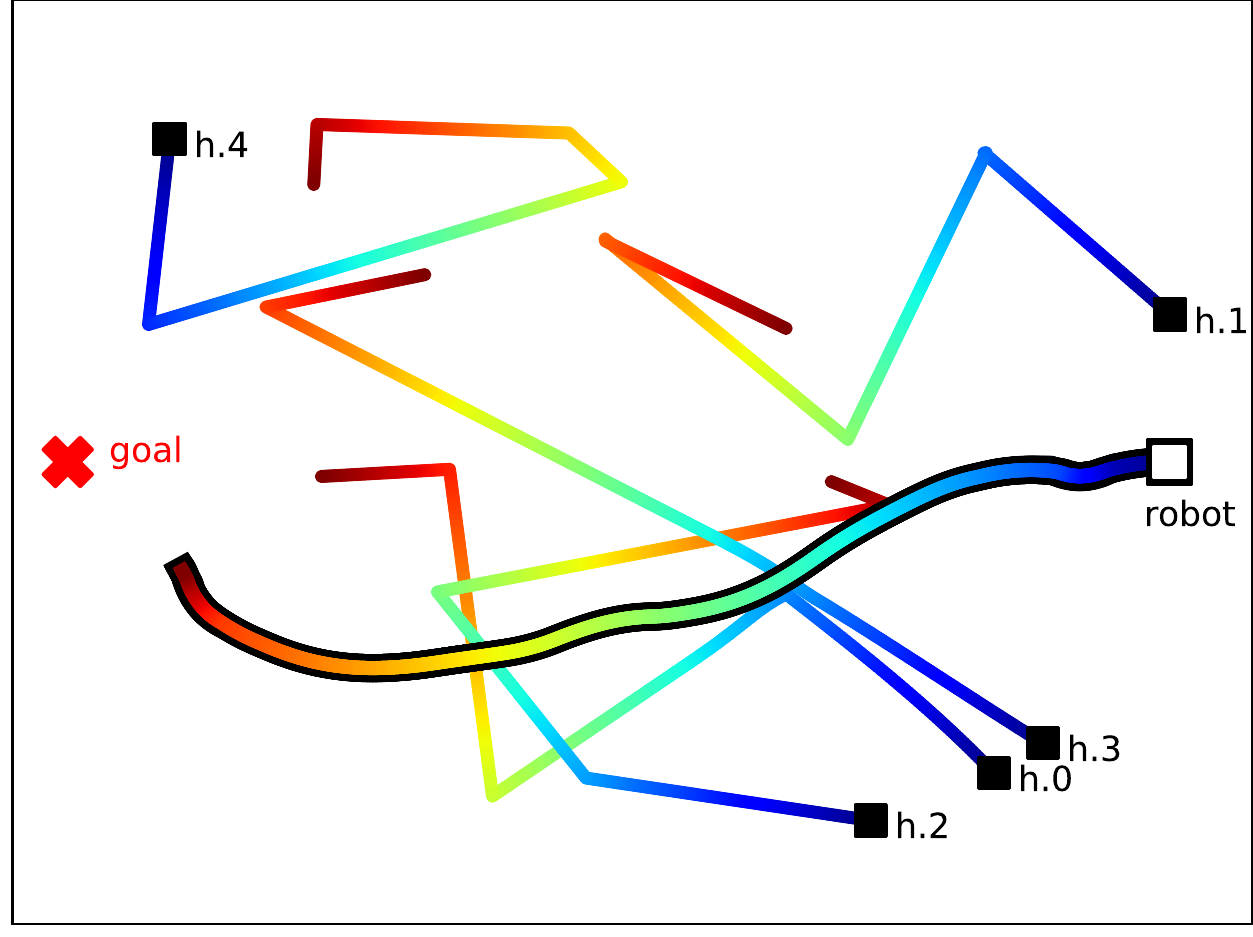}
  \caption{goal: 0.9, speed: 0.1 }
  \label{energydetPSK}
\end{subfigure}

\medskip 
\begin{subfigure}{.475\linewidth}
  \includegraphics[width=\linewidth]{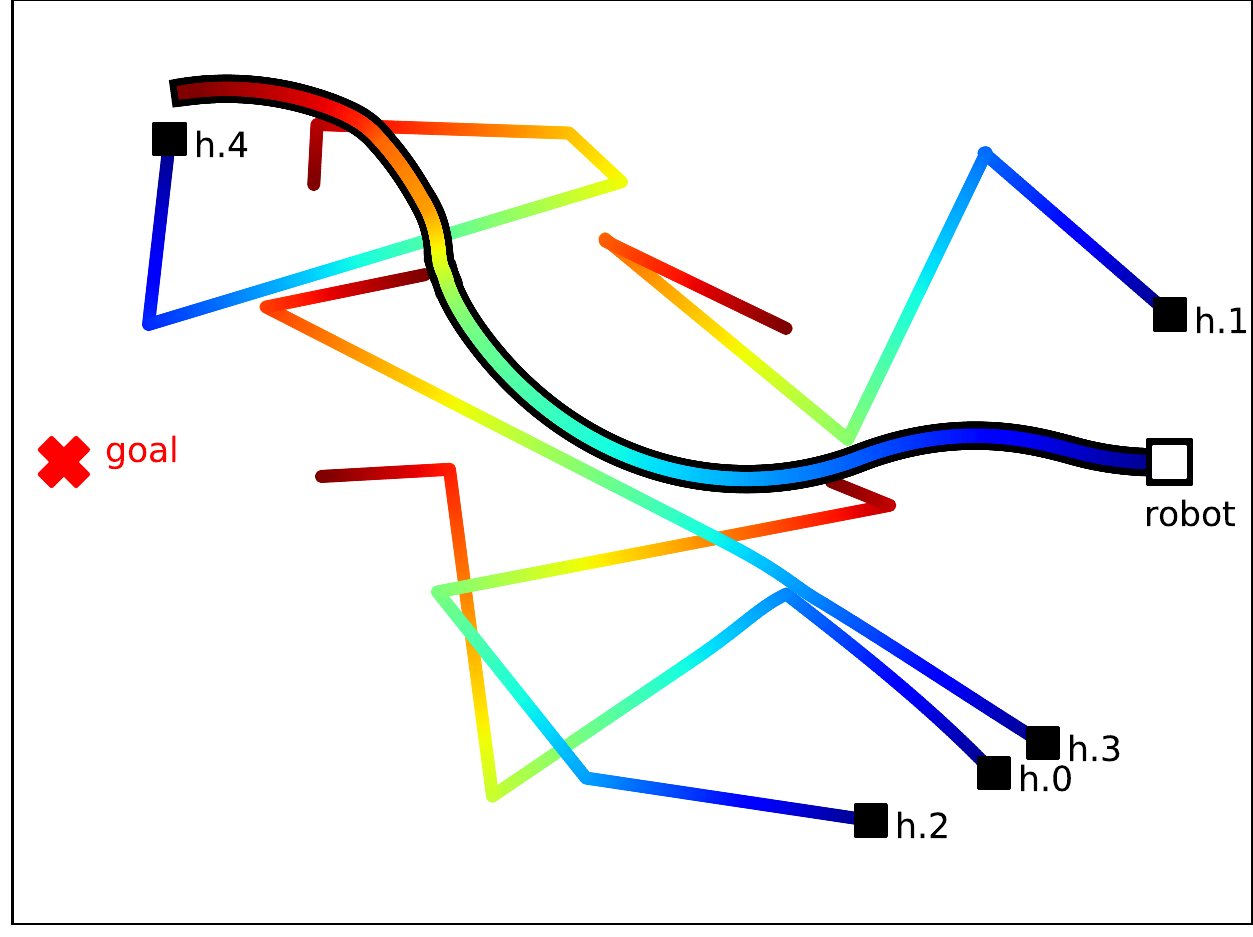}
  \caption{goal: 0.75, appr.: 0.25 }
  \label{velcomp}
\end{subfigure}\hfill 
\begin{subfigure}{.475\linewidth}
  \includegraphics[width=\linewidth]{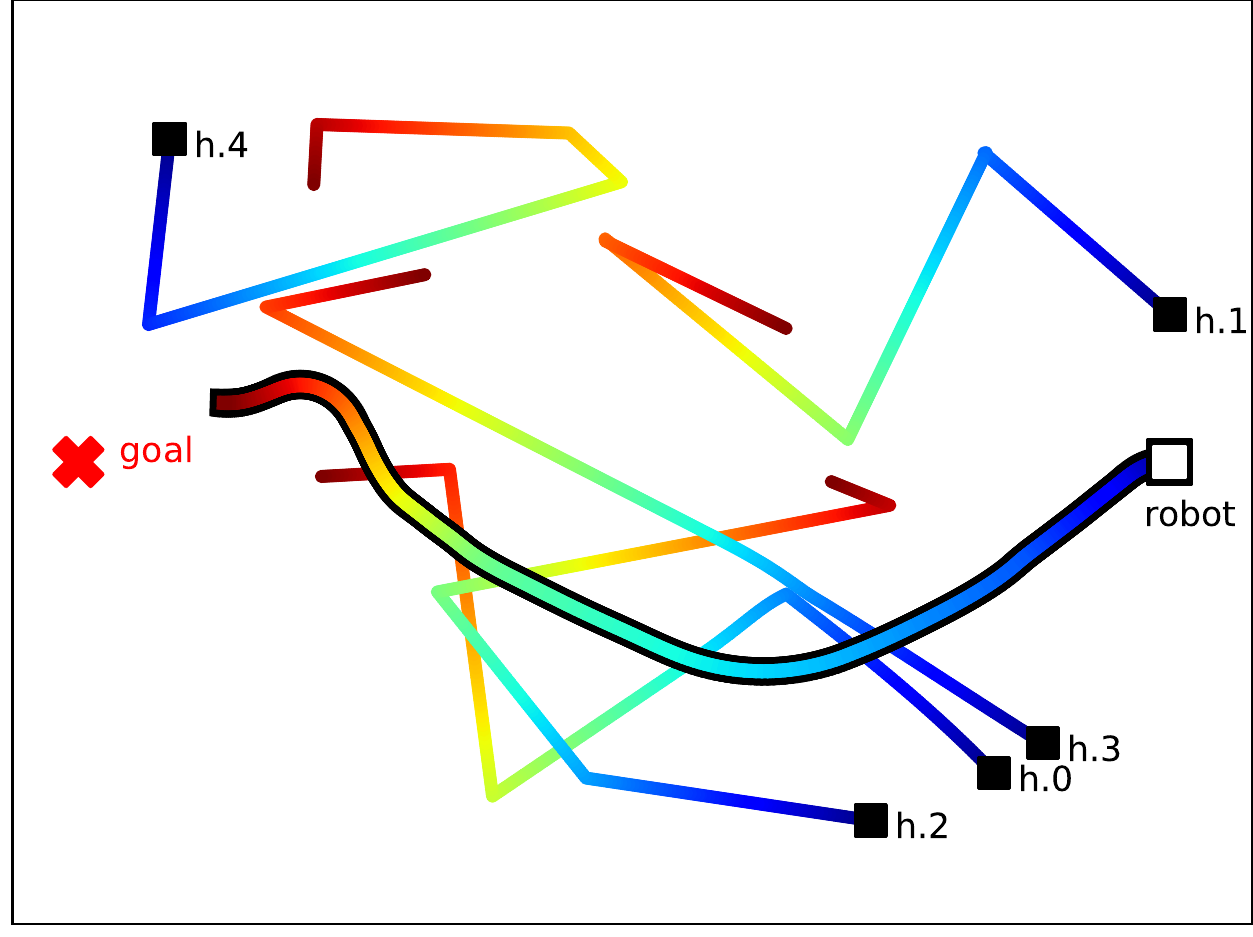}
  \caption{goal: 0.75, social: 0.25}
  \label{estcomp}
\end{subfigure}
\caption{\textbf{Trajectories in the social navigation environment for 4 different weight combinations of the reward function}. 
RBF-PEARL learns to generate different behaviors depending on the reward weight combination. 
The robot's trajectory is represented by the line with the black contour.
The color indicates the time step.}
\label{fig:social_behavior}
\end{figure}

We discuss six questions that we believe deserve some attention.
First, whether variational meta-RL is well suited to social robotics.
Second, what the difference is between the off-policy vs.\ on-policy approaches.
Third, the impact of the RBF layer in variational meta-RL.
Fourth, how the trainability of the RBF layer's parameters and the number of RBF neurons influence its performance.
Fifth, what mechanisms are behind the improved performance of the RBF layer.
And last, open research directions in the field of meta-RL for social robotics.

\begin{table}[b]
    \centering
\caption{\textbf{Final performance of PEARL, RBF-PEARL and on-policy baselines}
RBF-PEARL outperforms PEARL and the learning of RBF parameters is in most environments beneficial over fixed parameters (RBF-PEARL-fp). On-policy algorithms (MAML and RL2) struggle to learn good policies with the same number of environment steps.
Reported are the mean $\pm$ standard deviation over 5 seeds for each algorithm meta-testing performance at the end of meta-training.
}
\label{tab:ablation-study}
\resizebox{\columnwidth}{!}{
\begin{tabular}{lcccc}
 \toprule
 Algorithm & Lin.~Gaze & Non-Lin.~Gaze & Social Nav. & Racer\\
 \midrule
 PEARL &   $516 \pm10$  & $460 \pm16$  &  $297 \pm30$ & $56 \pm 6$\\
 Vanilla PEARL   &   $514 \pm 6$  & $469 \pm 16$   &  $253 \pm18$ & $56 \pm 5 $ \\
 RBF-PEARL &   $\textbf{549} \pm5$  & $\textbf{542} \pm10$  &  $\textbf{344} \pm31$ & $65 \pm 3$ \\
 RBF-PEARL-fp &   $543 \pm7$  & $518 \pm15$ &  $322 \pm12$ & $\textbf{72} \pm 6$\\
 MAML & $-99 \pm61$  & $203 \pm189$ & $-7 \pm97$  & $51 \pm 1$\\
 RL2 & $-89 \pm90$  & $-90 \pm81$ & $133 \pm74$  & $48 \pm 2$\\
 \bottomrule
\end{tabular}}
\end{table}

\subsection{Variational Meta-RL for Social Robotics}

We proposed the application of variational meta-RL to allow robots to quickly adapt to different social scenarios.
We achieved this by enabling robots to adapt to new reward functions which define the requirements of social scenarios, such as humans' different preferred social distances.
Indeed, our results from four simulation experiments show that with a variational meta-RL procedure (PEARL), robotic agents are able to quickly adapt to different scenarios, i.e., reward functions, requiring only 200 observations (Fig.~\ref{fig:gaze_results},~\ref{fig:social_nav_results},~and~\ref{fig:racer_results}).
Conversely, a classical RL algorithm trained on 200 observations (SAC 200) achieves a significantly lower performance. 
In the linear and non-linear gaze control environments, the performances of SAC 200 were respectively 23\% lower and 25 \% lower than those of RBF-PEARL. In the social navigation environment, the performances of SAC 200 were 47\% lower.
In the racer environment, the performances of SAC 200 were 53\% lower than those of RBF-PEARL.

To further demonstrate the interest of variational meta-RL, we compared the performance of SAC trained on more than 200 steps with the final models obtained with PEARL and RBF-PEARL.
As in the previous experiments, the two meta-RL algorithms only need 200 environment steps to adapt to each test environment.
Experiments were carried out for the linear-gaze, non-linear gaze, and racer environments (Fig.~\ref{fig:sac_vs_metarl}).
It should be noted that for these experiments, the number of gradient descent iterations per collected environment observation is the same for the meta-RL algorithms and for SAC.
In the previous experiments (Sec.~\ref{sec:result}), SAC was given more gradient descent iterations per observation to allow it to converge to a policy.
In the first two environments, SAC requires $50,000$ environment steps to reach the performance of PEARL and RBF-PEARL, which only benefit from $200$ environment steps.
Surprisingly, even after training for $200,000$ steps, SAC does not significantly outperform the two meta-RL algorithms.
This is especially true for the third environment (racer), where the performance of SAC does not seem to be superior to the meta-RL algorithms, even after $100,000$ environment steps.
It would appear that SAC does not manage to learn a moderately optimal action policy in the racer environment.
After looking carefully at our results, we realize that the optimal hyperparameters of SAC depend strongly on the learned task, and no set of parameters seems to be commonly optimal for all tasks.

Lastly, we evaluated whether the meta-policy learned by RBF-PEARL is able to produce diverse and meaningful behavior when adapted to different reward functions.  
We plotted the trajectories of the robot agent in the social navigation environment for four different reward weight combinations (Fig.~\ref{fig:social_behavior}). 
The meta-policy is adapted to each combination based on observations from 200 time steps.
The adapted behaviors are easily distinguishable from each other.
For a reward function that depends only on reaching the goal (a), the robot has trouble reaching the target position as it bumps into humans twice, and does not manage to modify its path accordingly.
However, weighting the speed component (b) more helps the robot to reach the target position, as the robot learns to have better control over its angular and linear speed.
By weighting the social and approach components (c and d) more, the robot actively tries to avoid people around it, even if this results in a failure to reach the target position.
In conclusion, variational meta-RL successfully and quickly adapts an agent to different reward functions, allowing it to efficiently find appropriate behaviors for different social scenarios.

\subsection{Sample efficiency of off-policy meta-RL}

We compared the performances of off-policy meta-RL, such as PEARL, with other on-policy methods proposed in the literature, such as MAML and RL2. 
We find that PEARL significantly outperforms on-policy meta-RL methods across all domains. 
PEARL converges to its final asymptotic performance with 3 to 10 times fewer samples during meta-training than other approaches proposed in the literature. 
Increasing the sample efficiency is generally speaking positive, specially for robotics applications. 
Indeed, running large amounts of environment steps in robotics applications is time-consuming, expensive, and requires significant human efforts. 
Fewer samples reduce the cost and effort required for data collection and labeling, making it more feasible to develop and deploy robotics systems in the real world. 

However, the actor-critic architecture used in models such as PEARL is associated with higher training complexity compared to models like MAML and RL2, as it requires the tuning of a greater number of hyperparameters. 
Also, PEARL may face more difficulties in generalizing to unseen environments, as the embedding may fail to capture the relevant information required for effective adaptation. 
Our proposed approach, RBF-PEARL, tends to mitigate this latter issue by transforming the task representation with our radial basis function (RBF) layer, and to improve performances compared to PEARL.

\subsection{Performance of PEARL vs.\ RBF-PEARL}

We compared the performance of RBF-PEARL to two versions of PEARL.
The first version, called PEARL, uses a neural network model with a similar number of parameters to RBF-PEARL.
This is achieved by using an MLP layer with $kd$ output neurons, where $k$ is the number of neurons of the RBF layer and $d$ is the latent dimension, followed by a ReLU activation unit 
instead of an RBF layer before the actor and critic network to process the task representation $z$.
All four experiments show that RBF-PEARL consistently outperforms PEARL in meta-test task performance (Fig.~\ref{fig:gaze_results},~\ref{fig:social_nav_results},~and~\ref{fig:racer_results}).
We further compared their asymptotic performances in the 20 meta-testing tasks in more detail (Table~\ref{tab:ablation-study}).
In the two linear environments, RBF-PEARL outperforms PEARL by a margin of 6\% in the linear gaze control environment and 16\% in the social navigation environment.
For the two non-linear environments the improvement is larger.
In the racer environment, the performance is 16\% better than that obtained by PEARL.
In the non-linear gaze control environment, the difference is 18\% compared to PEARL.

In addition, we report results obtained with a version of PEARL without the additional MLP layer, called Vanilla PEARL.
We find that the performances of PEARL and Vanilla PEARL are very similar in three of the four tested environments (Table~\ref{tab:ablation-study}).
The difference in the average final asymptotic return between the two is less than 5\%, and is within their confidence ranges.
PEARL is only significantly better than Vanilla PEARL (by 18\%) in the social navigation environment.

\begin{figure*}[t]

\centering

\includegraphics[width=.32\textwidth]{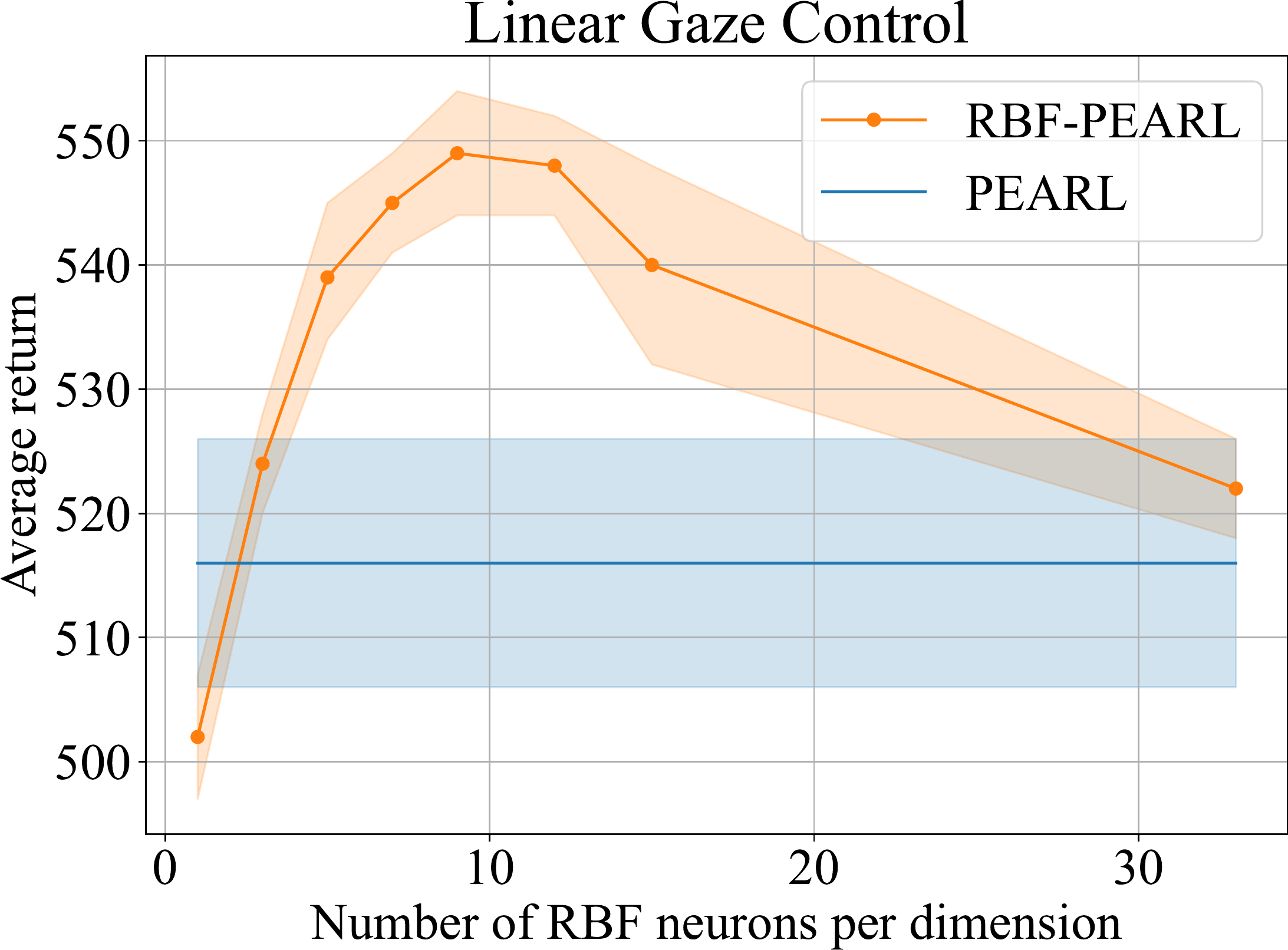}\hfill
\includegraphics[width=.32\textwidth]{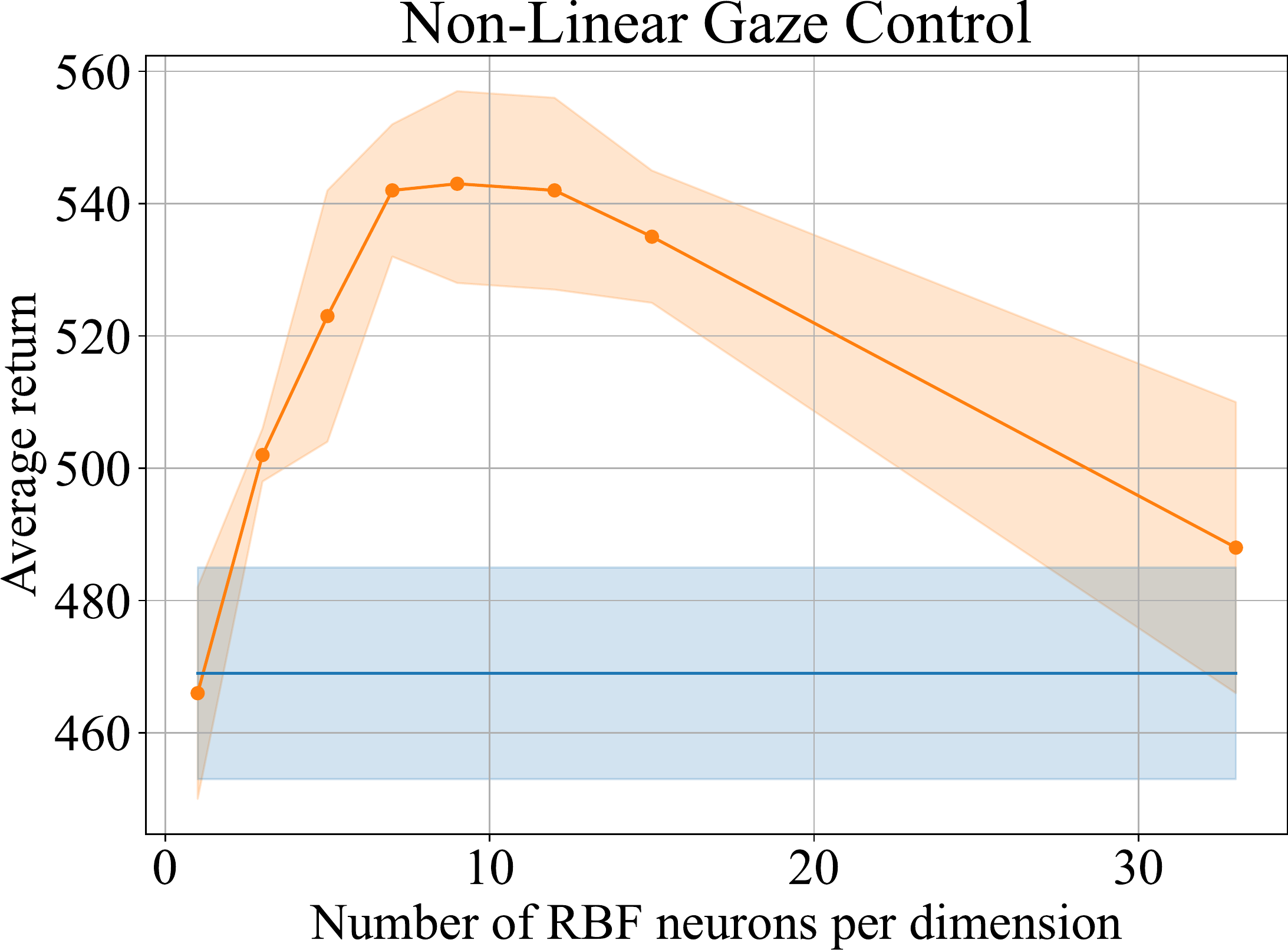}
\includegraphics[width=.32\textwidth]{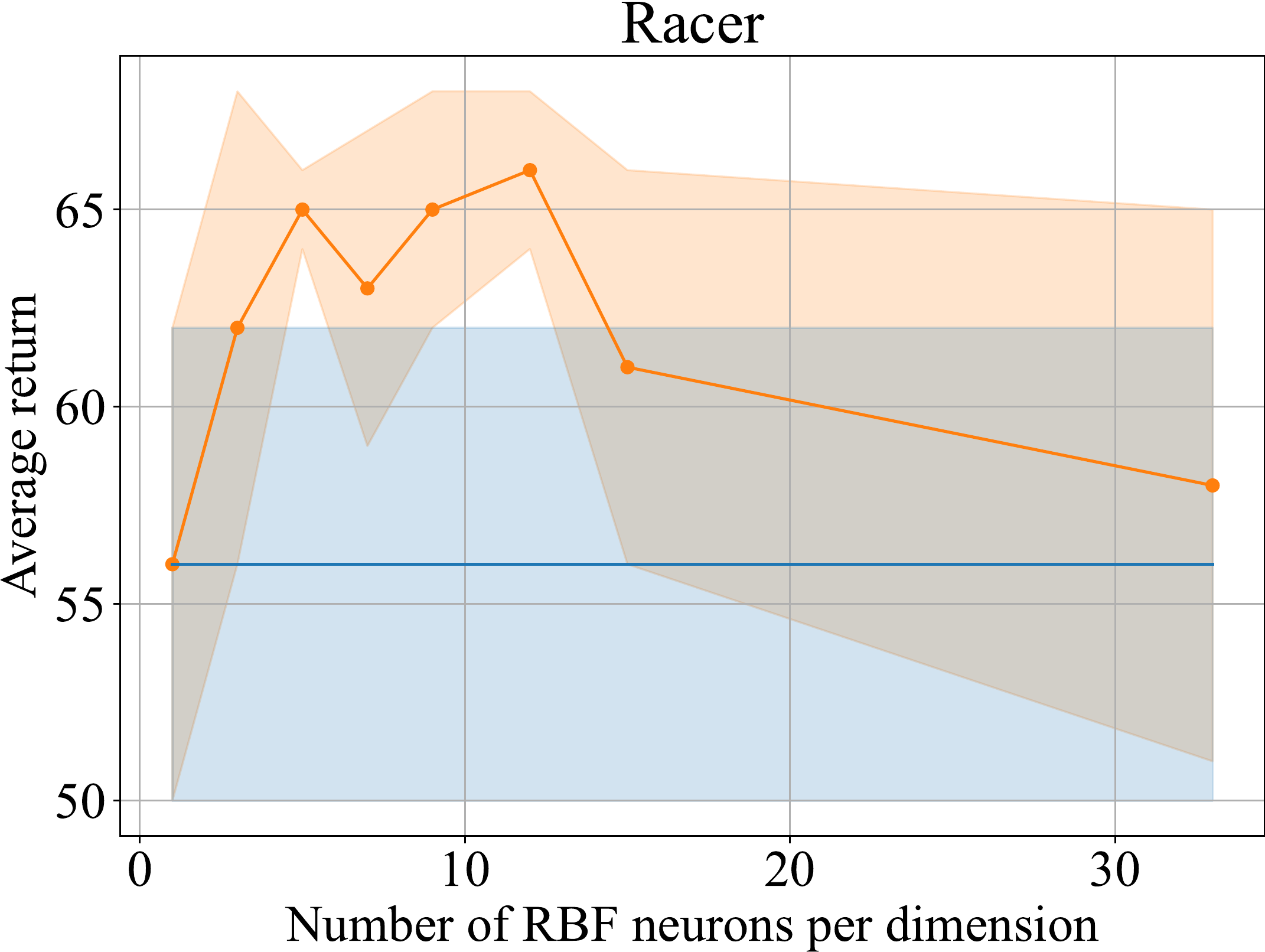}\hfill
    \caption{\textbf{Ablation study on the number of RBF neurons per input dimension for RBF-PEARL:}
    The optimal number of neurons is around 10 for the three evaluated environments.
    Reported are the  mean  and  standard  deviation over 5 seeds for each algorithm meta-testing performance after a certain number of meta-training steps.
    }
\label{fig:figure11}
\end{figure*}

\begin{figure*}[t]

\centering

\includegraphics[width=.32\textwidth]{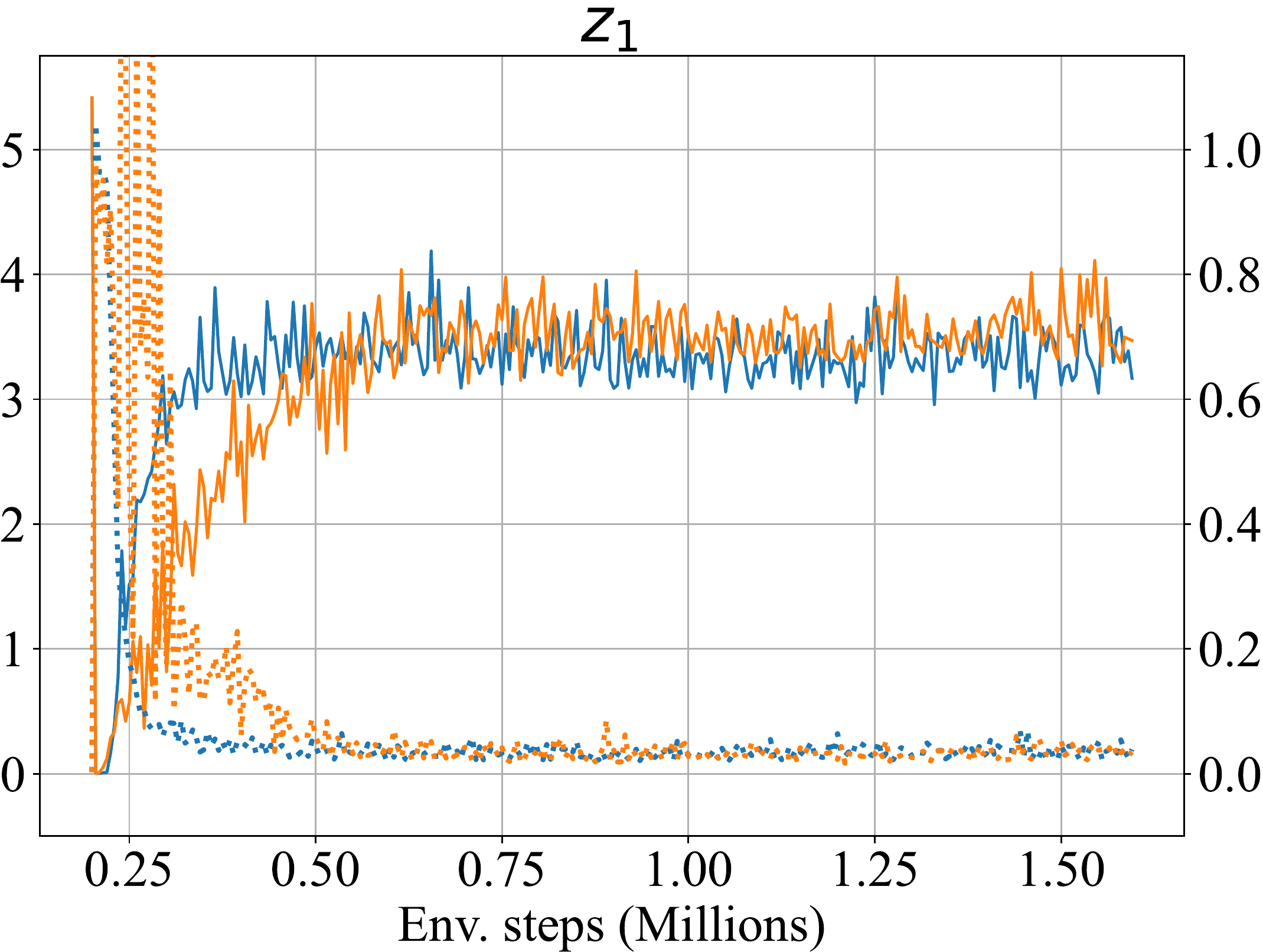}\hfill
\includegraphics[width=.32\textwidth]{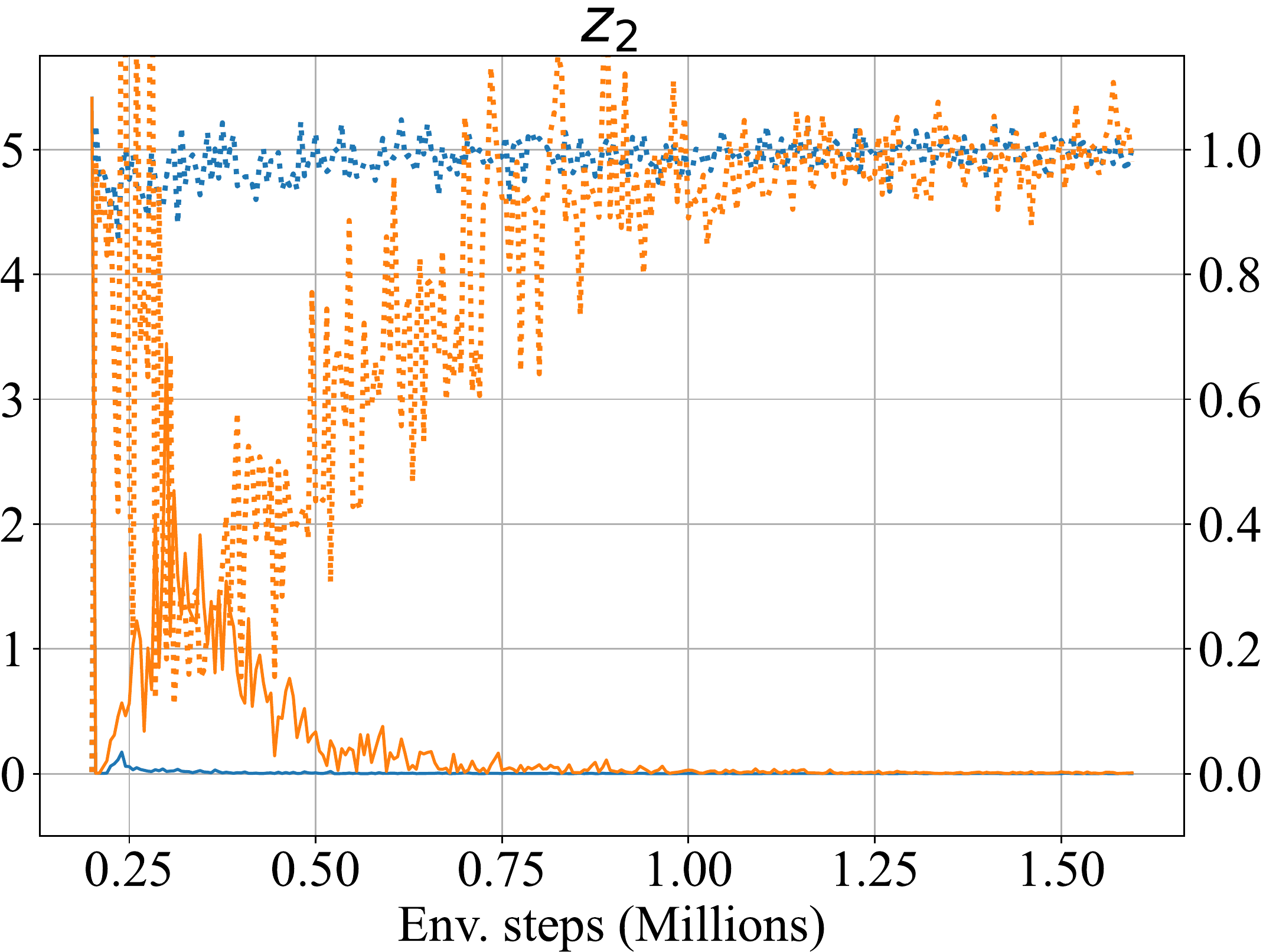}\hfill
\includegraphics[width=.32\textwidth]{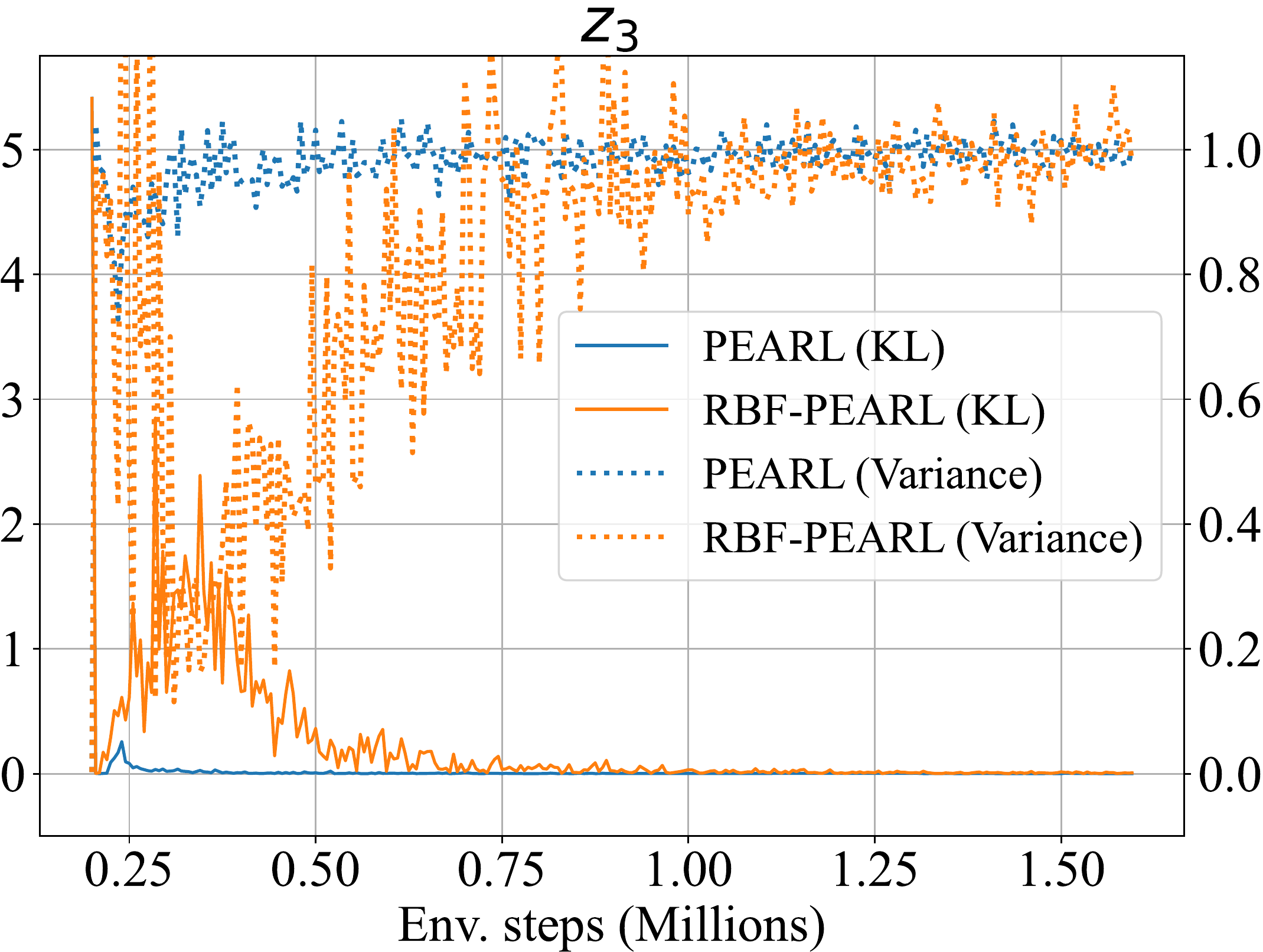}

\caption{\textbf{Kullback-Leibler (KL) divergence and variance of task representation $z$ for the linear gaze control environment:}
 Posterior collapse occurs in two dimensions ($z_2$, $z_3$) for both PEARL and RBF-PEARL (note that RBF-PEARL delays the collapse). We report the average KL divergence (solid lines) and variance (dotted lines) of each dimension of the task representation variable $z$ during meta-training. 
 The average is taken over five tasks chosen randomly among the 100 training tasks.
 }
\label{fig:figure9}

\end{figure*}

In summary, the results of PEARL and Vanilla PEARL compared to RBF-PEARL show that the RBF layer improves the performance of PEARL significantly.
This effect cannot be explained by a difference in their model capacity, as PEARL and RBF-PEARL have a similar number of parameters.
Instead, the computational properties of the RBF layer seem to be the important factors.

\subsection{Impact of trainability and number of RBF neurons}
We evaluated the effect of training the RBF parameters, i.e.,\ centers $c$ and scaling factors $\delta$ \eqref{rbf-eq}, to an RBF-PEARL architecture with fixed parameters (RBF-PEARL-fp).
The centers are fixed by evenly distributing them over an interval that was set to encompass the space of task representations $z$.
The scaling factors are fixed based on a function of the distance between center points.
They were chosen so that two neighboring RBF neurons both have an activation of $0.5$ for a representation $z_k \in \mathbb{R}$ lying in the middle between their centers.
Overall, we see that training the centers and scaling factors of the RBF layer has a beneficial impact on the asymptotic performances of the RBF-PEARL algorithm (Table~\ref{tab:ablation-study}).
In the non-linear gaze control and social navigation environments, training the RBF parameters improves the final asymptotic performances by 2\% to 6\%.
Only in the racer environment does the fact of having fixed parameters improve performance - by 10\%.
In summary, the advantage of learning the parameters of the RBF layer is task-dependent, but seems to be beneficial to most tasks.

Lastly, we analyze the effect of the number of neurons per input dimension in the RBF layer (Fig.~\ref{fig:figure11}).
In the three evaluated environments, we found that the number of neurons has a noticeable effect on the asymptotic performance.
The optimal number is around 10 in the three evaluated environments: 9 for linear gaze control, 9 for non-linear gaze control, and 12 for the racer.
We believe that the drop in performance for higher numbers of neurons may be due to overfitting on the training tasks. 

\subsection{How does the RBF layer improve performance?}

The RBF layer improves the performance of the variational meta-RL procedure, but what are the mechanisms behind this improvement?
In general, the layer can have three potential influences on the learning procedure.
First, as its input, task representation $z$ is also learned, and could alter the learning objective of $z$, resulting in a different representation.
Second, it could alter the temporal learning dynamics of the task representation, also leading to different learning dynamics in the downstream policy and value networks.
Third, its output $\tilde{z}$ could provide an improved representation of the policy and value networks.
We investigated these factors in the linear gaze control task (Sec.~\ref{c:gaze_control_env}).
We restricted our analysis to an RBF layer with the 3 RBF neurons per input dimension.
This low-dimensional representation makes it easier to analyze and visualize the results compared to the optimal configuration with the 9 RBF neurons per input dimension.

\paragraph{Influence on the learned input task representation $z$}

The learned task representations $z$ of PEARL without (Fig.~\ref{fig:posterior_collapse}, left-bottom) and with an RBF layer (Fig.~\ref{fig:figure10}) have only minor differences.
The average and standard deviation over the 100 meta-training task representation mean that $\mu$ per dimension are for PEARL: $z_1$: $0.041 \pm 3.676$ 
, $z_2$: $1.013 \pm 0.006$, $z_3$: $1.009 \pm 0.003$;
and for RBF-PEARL:
$z_1$: $0.061 \pm 3.507$
, $z_2$: $1.018 \pm 0.009$, $z_3$: $1.033 \pm 0.032$.

Both representations have a posterior collapse in two ($z_2$, $z_3$) of the three dimensions.
Only dimension $z_1$ represents a meaningful distinction of tasks.
The representation by PEARL without an RBF layer shows a minor larger spread of the task representations in dimension $z_1$ than RBF-PEARL ($3.676$ compared to $3.507$).
And RBF-PEARL has a minor larger spread in dimension $z_3$ ($0.032$ compared to $0.003$), but both these differences seem negligible.

Both representations show a clear clustering of tasks where tasks with a high reward weight on the visual component (red colored) are on one side in $z_1$.
Representations of the tasks with a high weight on the movement component are clustered at the opposite end (green colored).
Tasks with a high weight on the audio component (blue/brown colored) are in the middle.
Although the representations (with and without the RBF layer) are inverted to each other, both have this general cluster topology which should therefore not result in a difference in the downstream networks that learn based on them.

\begin{figure}[t] 
    \centering
    \includegraphics[width=.45\textwidth]{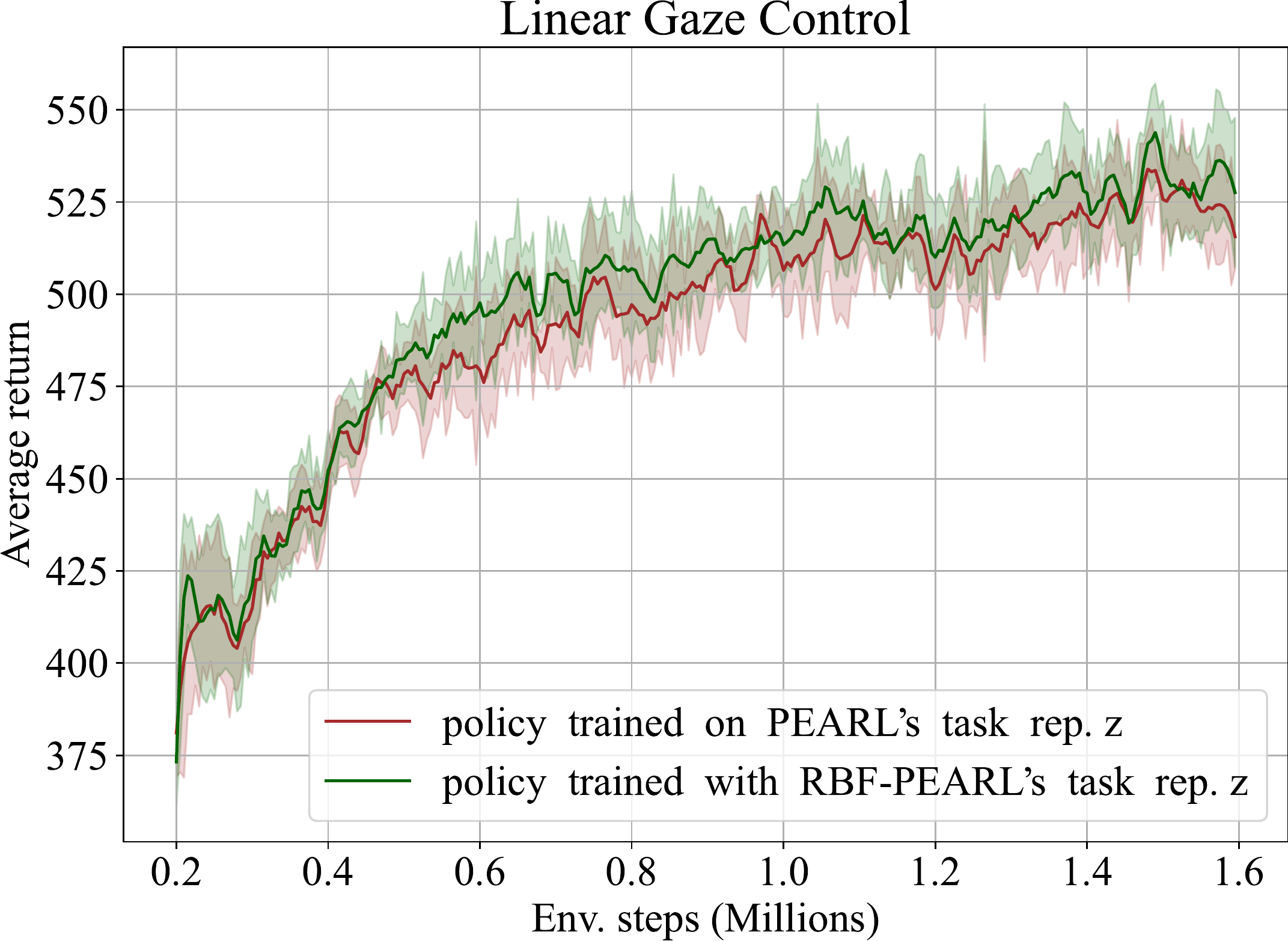} 
    \caption{\textbf{Performance of policies trained on learned (and frozen) context-encoders from PEARL and RBF-PEARL:} Test-task performance during meta-training on the linear gaze-control environment.
    It shows no significant differences between the two encoders, indicating that the performance difference between PEARL and RBF-PEARL is not due to their differences in the learned task representation $z$.
    } 
    \label{fig:context_encoder} 
\end{figure}

\begin{figure*}[t!]
	\centering
	\setlength\tabcolsep{0pt}
	\begin{tabular}{@{}ccc@{}}
        \small Task Representation $z$  & \small RBF Gaussians for $z_1$  & \small RBF Task Representation $\tilde{z}_1$ 
        \\
        \includegraphics[trim={0.2cm 0 0 3cm},clip,width=0.26\textwidth]{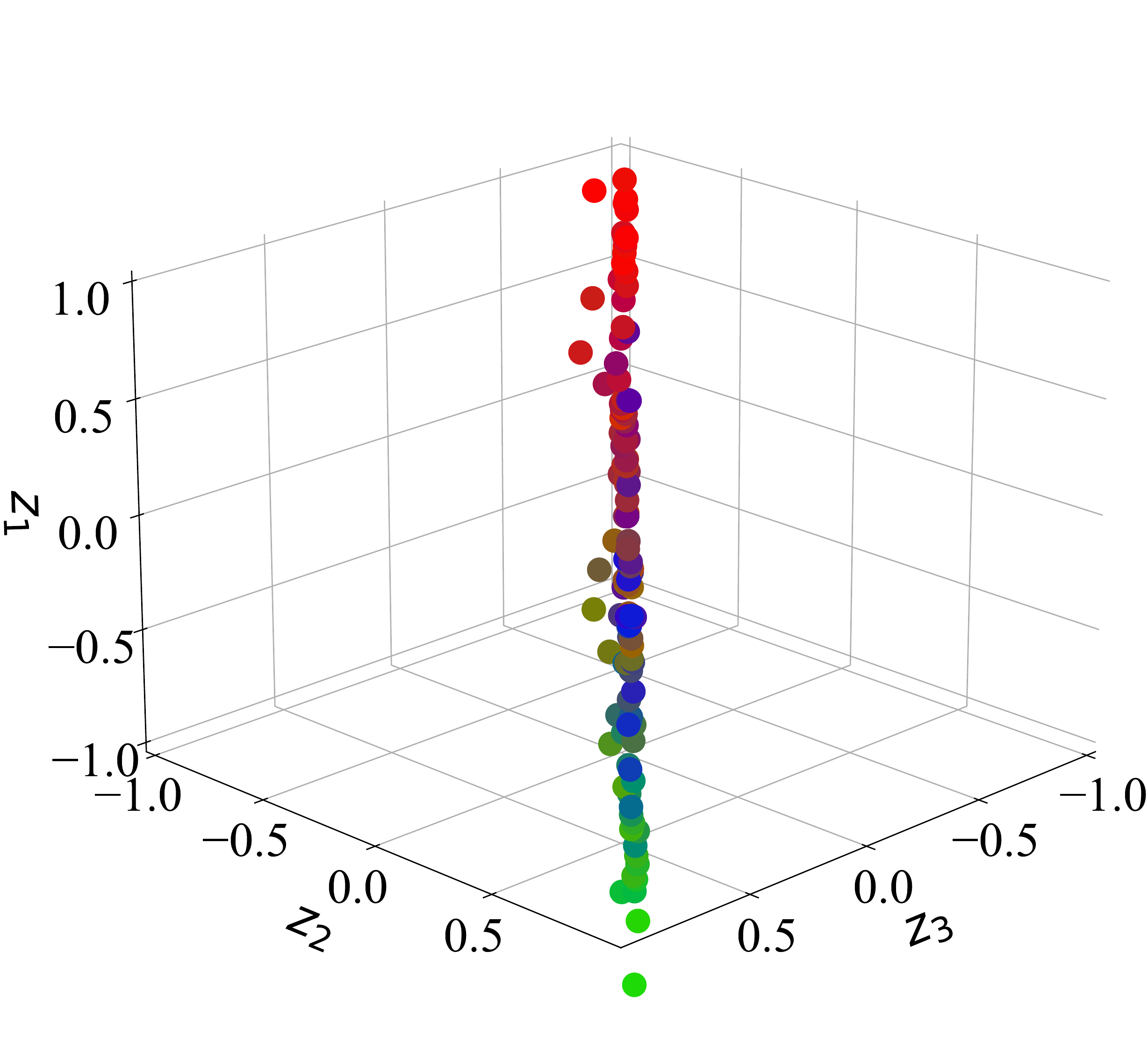} & ~~
	    \includegraphics[width=0.4\textwidth]{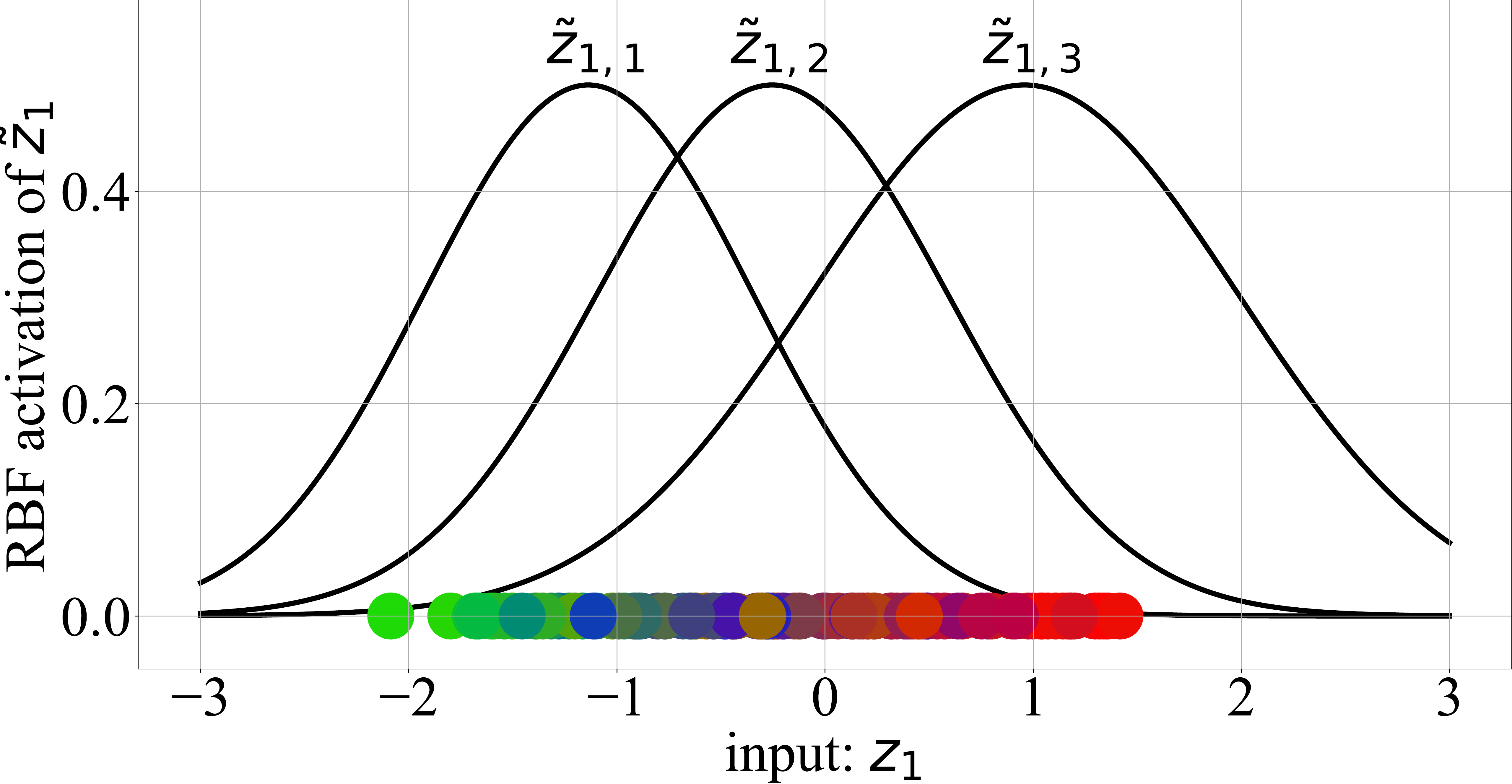} ~~&
	    \includegraphics[trim={0 0 0 3cm},clip,width=0.26\textwidth]{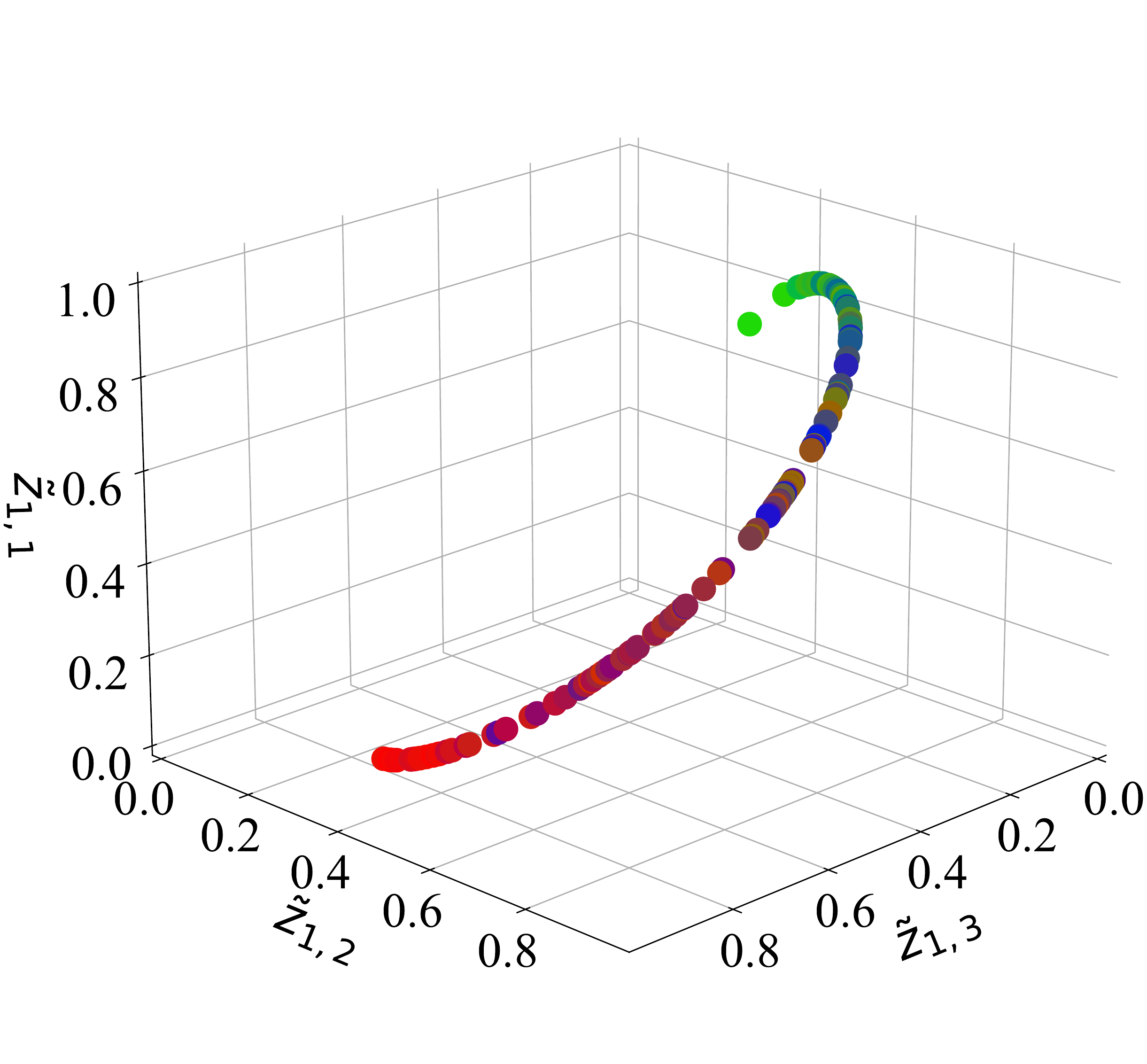}
        \\
	\end{tabular}	
	\caption
	{\textbf{RBF-PEARL's task representation with the Gaussian associated with dimension $z_1$ with no posterior collapse:}
The RBF layer projects task representation $z_1 \in \mathbb{R}$ in a higher dimension $\tilde{z}_1 \in \mathbb{R}^3$,
where each RBF Gaussian learns to represent a specific task type.
Each colored dot corresponds to the representation of one of the 100 meta-training tasks.
The color represents the predominant reward weight component of the reward function, i.e.,\ the task type.
Red: largest weight is the visual weight.
Blue/brown: high audio weights.
Green: high movement weights.
(left): Three-dimensional task representation $z$ learned by the task encoder.
The tasks are only spread in dimension $z_1$.
The other dimensions ($z_2, z_3$) have a posterior collapse.
(middle): RBF Gaussians associated with the input dimension without posterior collapse ($z_1$).
Each Gaussian specializes in order to represent a different task type.
(right): The three-dimensional representation obtained by the RBF layer for input dimension $z_1$.}
	\label{fig:figure10}
\end{figure*}

We further evaluated whether the learned representation $z$ obtained with the RBF layer has an influence on the performance increase.
We trained an actor and critic network with the pre-trained context encoder obtained from PEARL and RBF-PEARL.
No significant differences, either in their learning curves or in their final performances, can be observed (Fig.~\ref{fig:context_encoder}). 
This indicates that differences in the performances of PEARL vs.\ RBF-PEARL do not result from their differences in the learned representation $z$.

In summary, the differences between the learned task representation $z$ with and without the RBF layer are minor.
They do not explain the increase in the performance of RBF-PEARL.

\paragraph{Influence on the temporal dynamics of learning task representation $z$}

We analyzed the temporal dynamics of learning $z$ by looking at how posterior collapse happens on the three dimensions of task representation $z$.
The differences in how posterior collapse occurred between PEARL and RBF-PEARL could explain their performance differences.
To measure this, we examine the KL divergence and the variance of the posterior distribution on each of the dimensions of the task representation during the meta-training stage (Fig.~\ref{fig:figure9}).
Posterior collapse occurs for both methods in two of the three dimensions ($z_2$, $z_3$).
Nonetheless, RBF-PEARL delays the posterior collapse for 400k steps compared to PEARL.
Similarly, for dimension $z_1$, RBF-PEARL requires more time to learn the final representation as shown by the longer time of the KL loss and variance to reach their asymptotic levels (Fig.~\ref{fig:figure9}, left).
This delay could explain why RBF-PEARL performs slightly below PEARL for the first 400k steps (Fig.~\ref{fig:gaze_results}, left).
Afterward, the KL loss and variance are similar between RBF-PEARL and PEARL.

In conclusion, the RBF layer has a temporal effect on the learning of task representation $z$ by delaying it.
This includes a delay in the posterior collapse.
This might affect the final performance of the RL algorithm, for example by inducing a greater exploration during learning.
Nonetheless, the impact of this effect cannot be clearly defined and we believe it to be of minor consequence to the learning performance.

\paragraph{Influence of the output representation $\tilde{z}$}

The final influence that the RBF layer has on the performance of RBF-PEARL is through its output representation $\tilde{z}$, which is given as an input to the downstream policy and value networks instead of $z$.
We visualized this representation for the RBF neurons that encode the non-collapsing dimension $z_1$ (Fig.~\ref{fig:figure10}, right).
It lifts the one-dimensional representation $z_1$ into a three-dimensional space $\tilde{z}_{1}$.
Analyzing the shape of the Gaussians associated with $\tilde{z}_{1}$ (Fig.~\ref{fig:figure10}, middle), each Gaussian is centered around a specific cluster of task representations.
The first Gaussian $\tilde{z}_{1,1}$ specializes in tasks with a high weight on the movement component (green colored).
Representations of tasks with a high weight on the audio component (blue/brown colored) are centered around the activation region of the second Gaussian $\tilde{z}_{1,2}$.
The third Gaussian $\tilde{z}_{1,3}$ specializes in tasks with a high weight on the visual component (red colored).
We believe that this effect is the main cause of the RBF layer's performance increase.
The objective of the downstream networks is to learn specific policies and value functions for the different tasks, i.e.,\ tasks in which the visual, audio, or movement component is more important.
Differentiating between these tasks is difficult from the one-dimensional representation $z_1$ learned by the standard PEARL.
For example, to identify audio tasks which are clustered in the middle of the representation in $z_1$ (Fig.~\ref{fig:posterior_collapse}, left-bottom), the downstream networks have to learn a rule that defines this region using two borders: $y > z_1 > x$.
In contrast, for RBF representation $\tilde{z}_1$ it is only necessary to identify whether a certain Gaussian has a large activation.
In the case of audio tasks, the second Gaussian should be mainly activated: $\tilde{z}_{1,2} > x$.
This seems to reduce the complexity of the rules that the downstream networks have to learn to identify tasks, making it easier to learn specific policies and values for them.

In summary, we believe that the main effect that the RBF layer has in improving performances is based on its changed task representation $\tilde{z}$.
The representation seems to allow the downstream networks to identify certain tasks more easily and to learn specific outputs for them.
Nonetheless, this explanation is only an intuition and should be explored further in future research.

\subsection{Open research directions}

Meta-RL for social robotics is currently underexplored and several research directions are still open.
This section discusses some of these directions.

\paragraph{Partially observable environments} 
In social robotics, it is often unrealistic to assume that the agent has complete information. 
Factors such as sensor noise, occlusions, and limited field of view can all contribute to incomplete observation data. 
Incomplete observations can have a significant impact on the training performance of the agent, which can make it challenging to learn an effective policy for the task at hand~\cite{quintero2022human,rosano2021embodied}. 
Meta-reinforcement learning is a promising approach to tackle the problem of partially observable environments. 
By leveraging previous experience and learning how to learn, the agent can quickly adapt to new tasks and environments, even when observation information is incomplete or noisy. 
Meta-RL has already been studied in the context of adapting to unseen environments with a limited field of view~\cite{wortsman2019learning} or sim2real scenarios~\cite{arndt2020meta}, where the agent needs to generalize to new and diverse settings. 
In our paper, the gaze control environment can be considered a partially observable environment, as the visual information the robotic head can obtain is limited by the field of view of the camera. 
However, further research is needed to explore the effectiveness of off-policy meta-RL for partially observable environments in the specific context of social robotics.

\paragraph{Safe reinforcement learning} 
Safety in reinforcement learning can be particularly important in the context of social robotics, where robots are designed to interact with humans in shared environments. 
Social robots are expected to operate safely and interact with humans in a way that is consistent with social norms and values. 
While meta-RL in itself cannot guarantee that a robot will operate safely in social environments at test time, it is possible to integrate existing approaches from the literature into meta-RL algorithms. 
Some common approaches to safe reinforcement learning include adding constraint/regularization methods to the objective function~\cite{yang2021wcsac}, using ensemble networks to capture uncertainties in the environment~\cite{lutjens2019safe}, or using a hierarchical control architecture with a manually designed set of constraints to control the low-level policy~\cite{xiong2022hisarl}. 
Recent approaches also propose to use meta-RL in environments with non-stationary disturbances to adapt the safety constraints to the disturbances in the environment~\cite{chen2021context}. 
Nevertheless, further research is needed to integrate safety constraints into the meta-reinforcement learning framework.

\paragraph{Multi-agent reinforcement learning} 
In the context of social robotics, multi-agent reinforcement learning (MARL) has been proposed to address situations that involve several agents/robots~\cite{everett2018motion,semnani2020multi}. 
In MARL, each agent is represented by a learning algorithm that interacts with the environment and other agents to optimize its performance. 
The agents can be homogeneous, meaning that they are identical and have the same objective, or heterogeneous, meaning that they have different objectives and learning algorithms. 
The environment can be either cooperative, where agents work together to achieve a common goal, or competitive, where agents compete with each other to achieve their individual goals. 

One of the primary challenges in this setup is to coordinate several continuously learning and differently behaving agents.
The changing behavior of one agent can result in a negative outcome for another agent because it is not adapted to the new behavior. Social environments involve human participants, making the problem of generalization even more challenging. 
Some meta-RL approaches~\cite{charakorn2021learning,he2023learning} address this by assuming a distribution of agents available for practicing.
The resulting policy should then be able to generalize and adapt to environments with agents cooperating in different ways. 
A similar approach could be used for social robotics where not only are other agents considered for adaptation, but also humans.

\section{Conclusion}
\label{sec:conclusion}
In this exploratory study, we investigate the use and limitations of variational meta-RL for social robotics.
We show that meta-RL successfully learns to adapt quickly (within 200 steps) to a new reward function that describes a desired social behavior in a different environment. 
However, in our task state-of-the-art methods (PEARL) exhibited posterior collapse, which is problematic in meta-RL since the encoder is supposed to accumulate information for better generalization, and collapsed encoding dimensions cannot do so. 
We started investigating how to mitigate posterior collapse in variational meta-RL by adding an RBF network after each encoded dimension. 
Based on the result in Tab.~\ref{tab:ablation-study}, and the analysis of the representation learned by RBF-PEARL, our algorithm improves the performances of meta-RL algorithms for reward design in social robotics. 
Our RBF layer improves the asymptotic performances of the PEARL algorithm in several different problem domains, all inspired by social robotics tasks. The PEARL algorithm learns a sub-optimal representation of the task. 
While we do not solve this issue, the RBF-PEARL algorithm mitigates the effect of this sub-optimal representation, providing the actor and critic network with a different representation of the task using the dimensions of the task representation that do not suffer from posterior collapse.
The results clearly demonstrate that an RBF layer reduces the effect of posterior collapse, and allows for steeper learning curves and higher asymptotic performance. 
We believe that such studies pave the way to a better understanding of meta-RL for social robotics, a clearly underinvestigated domain. 
We hope that our findings will help foster research in this direction.

\section*{Declarations}

\paragraph{Data}
The authors declare that the data supporting the findings of this study are available within the article.

\paragraph{Code}
The code used during the current study is available from the corresponding author on reasonable request.

\paragraph{Funding}
Partial financial support came from the ANR MIAI institute (ANR-19-P3IA-0003), the Horizon 2020 SPRING project (\#871245), and from the ANR ML3RI (ANR-19-CE33-0008-01).
The authors have no competing interests to declare that are relevant to the content of this article.

\bibliography{mybibfile}

\end{document}